\documentclass[journal]{IEEEtran}
\usepackage{cite}
\usepackage{hyperref}       
\usepackage{url}
\usepackage{booktabs}       
\usepackage{amsfonts}    
\usepackage{amssymb}
\usepackage{amsmath}
\usepackage{graphicx}
\usepackage{caption}
\usepackage{algpseudocode, algorithmicx, algorithm,mismath}
\usepackage{siunitx}
\usepackage{multirow}
\usepackage{pdfpages}
\usepackage{array}

\newcommand \argmax {\operatorname*{argmax}}

\newcommand \andothers {\textit{et~al.}}
\ifCLASSINFOpdf
\else
\fi
\ifCLASSOPTIONcompsoc
  \usepackage[caption=false,font=normalsize,labelfont=sf,textfont=sf]{subfig}
\else
  \usepackage[caption=false,font=footnotesize]{subfig}
\fi

\ifCLASSOPTIONcaptionsoff
  \usepackage[nomarkers]{endfloat}
 \let\MYoriglatexcaption\caption
 \renewcommand{\caption}[2][\relax]{\MYoriglatexcaption[#2]{#2}}
\fi

\usepackage{url}

\begin{document}

\title{Anomaly Detection of Defect using Energy of Point Pattern Features within Random Finite Set Framework}

\author{\href{https://orcid.org/0000-0002-7441-9344}{\includegraphics[scale=0.06]{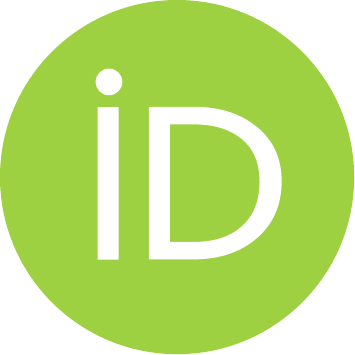}\hspace{1mm}Ammar Kamoona,~\IEEEmembership{}}
        \href{https://orcid.org/0000-0002-4800-6554}{\includegraphics[scale=0.06]{orcid.pdf}\hspace{1mm}Amirali Khodadadian Gostar},~\IEEEmembership{}
        \href{https://orcid.org/0000-0002-6192-2303}{\includegraphics[scale=0.06]{orcid.pdf}\hspace{1mm}Alireza Bab-Hadiashar} ~\IEEEmembership{},and~\href{https://orcid.org/0000-0001-9525-1467}{\includegraphics[scale=0.06]{orcid.pdf}\hspace{1mm}Reza Hoseinnezhad},~\IEEEmembership{}
\thanks{A. Kammona is with Engineering college, Royal Melbourne Institute of Technology, Melbounre,
Australia, e-mail: (s3442819@student.rmit.edu.au).}
\thanks{Manuscript received April 19, 2005; revised August 26, 2021.}}

%
%

\markboth{to be submitted to IEEE TRANSACTIONS ON INDUSTRIAL INFORMATICS,~}%
{Shell \MakeLowercase{\textit{et al.}}: Bare Demo of IEEEtran.cls for IEEE Journals}
%



\maketitle

\begin{abstract}
In this paper, we propose an efficient approach for industrial defect detection that is modeled based on anomaly detection using point pattern data. Most recent works use \textit{global features} for feature extraction to summarize image content. However, global features are not robust against lighting and viewpoint changes and do not describe the image's geometrical information to be fully utilized in the manufacturing industry. To the best of our knowledge, we are the first to propose using transfer learning of local/point pattern features to overcome these limitations and capture geometrical information of the image regions. We model these local/point pattern features as a random finite set (RFS). In addition we propose RFS energy, in contrast to RFS likelihood as anomaly score. The similarity distribution of point pattern features of  the normal sample has been modeled as a multivariate Gaussian. Parameters learning of the proposed RFS energy does not require any heavy computation. We evaluate the proposed approach on the MVTec AD dataset, a multi-object defect detection dataset. Experimental results show the outstanding performance of our proposed approach compared to the state-of-the-art methods, and the proposed RFS energy outperforms the state-of-the-art in the few shot learning settings. 
\end{abstract}


%
\IEEEpeerreviewmaketitle

\section{Introduction}
%
%
%
%
\IEEEPARstart{A}{utomated} visual inspection in the manufacturing process is highly important for quality management of the manufactured product. Poor quality products provoke cost increase in the handling defect within the warranty period and decay in market reputation. To avoid unnecessary losses and improve quality, automated visual inspection is a development trend in industrial intelligence~\cite{ngan2011automated}. In this context, defect detection is one of the critical challenges that need to be addressed, which paves the way for full industrial intelligence~\cite{gao2021review}.

The early approach of defect detection was carried out manually by an expert inspector. However, the efficiency is extremely low to satisfy the industrial intelligence, and recognition of defects is not stable due to that inspectors are prone to fatigue~\cite{neogi2014review}.

With the development of vision-based sensors, vision-based defect detection has attracted wide attention from both industry and academia. Vision-based defect detection employs computer vision techniques to detect a defect via a product's images, which provides a fast, economical and stable manner. Accordingly, it has been widely used in many fields, such as steel~\cite{wang2018simple}, wood~\cite{he2019fully}, ceramic~\cite{hanzaei2017automatic}, fabric~\cite{ngan2011automated}, and architecture~\cite{shen2013automated}.

Early approaches in this domain that use camera sensor data involve employing various image processing techniques. Several examples comprise the use of Haar filter for tile surface inspection~\cite{elbehiery2005visual}, the use of local order binary pattern for fabric defect detection~\cite{jing2013fabric}, and the use of SIFT features for PCB inspection~\cite{lowe2004distinctive}. Later, the use of both image processing and machine learning algorithms has also shown satisfactory performance. A simple examples of this is fabric defect detection via the use of the histogram of oriented gradients (HOG) features with a support vector machine (SVM)~\cite{shumin2011adaboost}.

The advance in machine learning, especially representation learning or feature learning, provide an alternative approach to overcome the drawbacks of using image processing techniques. The significant problem with image processing techniques is that they use implicit engineering features that fail to address complex scenes. A recent study done by Wang \andothers~\cite{wang2018deep} highlights the potential of using deep learning in smart manufacturing. In the light of defect detection, various deep learning techniques have been proposed for different surface defect detection~\cite{dai2020soldering,yin2020deep}.
Yu \andothers~\cite{yu2017fully} proposed to use a two-stage approach (segmentation and detection)
using fully convolutional neural network for surface defect inspection in
industrial scenes, and Wu \andothers~\cite{wu2017surface} proposed to use a CNN-based
general defect detection method, in which a multi-scale scheme
was used to obtain highly accurate identification. Wang~\andothers~\cite{wang2019lednet} proposed a LEDNet network for LED chips defect detection based on encoder-decoder networks. 

The effectiveness of deep representation learning, such as convolutional neural networks (CNNs), is constrained by the training samples' availability. This raises two problems: the class imbalance between the \textit{normal} and \textit{defected} samples and difficulty of data annotation. These two problems are well-known in the literature and a subject of continuing  research~\cite{hwang2019hexagan,oliver2018realistic,yoon2018gain}. In the manufacturing environment, it is quite a simple task to obtain normal samples as many as required. In contrast, it is extremely difficult or even possible to obtain a sufficient number of defective samples in a short period to train robust classification models for defect detection. Consequently, defect detection is conducted under unsupervised learning as anomaly detection or one class classification using normal samples only.  

Anomaly detection is defined as the process of identifying instances in the data that deviate from the predefined norm~\cite{chandola2009anomaly}. Anomaly detection in visual data, especially images, aims to find "anomalous" images with irregularities. Correspondingly, image anomaly detection poses a fundamental problem in computer vision and has various applications starting from quality control~\cite{bergmann2019mvtec} to medical images~\cite{baur2018deep}. The uniqueness of anomaly detection compared to supervised classification problem is in two folds: (i) the lack of anomalous samples either labeled or unlabeled comes due to the nature of the problem, and (ii) the small difference between normal and anomalous images that are often fine-grained as the anomalous area might be very small in high-resolution images. Due to the lack of anomalous samples, the anomaly detection problem is addressed by building an anomaly detector using only normal samples. One approach of anomaly detection is performed based on deviation from the statistical distribution of normal samples.

Bergmann~\textit{et~al}.~\cite{bergmann2019mvtec} proposed an encoder-decoder network for anomaly detection of defect.  In addition, to overcome the limitation of having a multi-objects defect detection dataset, they proposed an MVTec-AD dataset, which includes different objects and textures with a wide variety of defects. Different works in this domain have been proposed~\cite{bergmann2019mvtec,rudolph2021same}, and they concentrate on performing anomaly detection by looking at a whole image by extracting a single feature vector known as a global feature.  The global feature is very effective at summarizing the content of an entire image~\cite{cao2020unifying}. In contrast, local features are more suitable for describing the geometrical information regarding a specific part of the image. Local features have been widely used in different computer vision applications, such as SLAM, short for simultaneous localization and mapping~\cite{tang2019gcnv2}, Structure from Motion (SfM)~\cite{schonberger2018semantic}. These features offer very memory-efficient representation  and most important are more robust under extreme viewpoint changes and lighting conditions.

In defect detection, the aim is to detect different irregularities in the image, even the slight ones. This implies that using local features to capture these variations is more intuitive and practical since they are more robust against illumination and viewpoint changes. To the best of our knowledge, no work in the literature has explored the use of transfer learning of local features or \textit {point pattern features} for defect detection. The contributions of the paper are as follows: 
\begin{itemize}
	\item We propose using transfer learning of local features or point-pattern features instead of commonly used global feature-based methods. We model these features as unordered set within random finite set statistics.
	\item We propose to model the set of features using an energy-based model to explicitly represent the probability distribution of the random finite set (RFS) features.
	\item We propose to learn the parameters of the RFS energy on the high dimension to avoid the decoupling problem due to dimension reduction.
	\item As opposed to other methods, the proposed approach is unique. The uniqueness of the approach comes from the fact that it is very computationally efficient and requires few samples to train, proven by few-shot experiments.	
\end{itemize}

The rest of this paper is organized as follows: Section~\ref{Sec:relate_dworks} summaries the most current works on defect detection and point pattern feature extraction,  Section~\ref{Sec:background} provides background on the energy-based model and RFS framework and how it can be employed for anomaly detection. Section~\ref{Sec:proposed_appraoch} covers the proposed approach used in the paper. Section~\ref{Sec:experiments} presents experimental results, and finally Section~\ref{Sec:conclusion} concludes the paper.

\section{Related works}
\label{Sec:relate_dworks}
In this section, we review the most recent works on defect detection. The existing anomaly detection of defects can be roughly classified into two approaches, generative-based models and transfer learning-based models (pre-trained networks). The focus of this review is on anomaly detection rather than localization. In addition, we review point-pattern-based feature extraction methods since this work is the first to use these features for defect detection.

\subsection{Defect Detection with generative models}
Generative models \cite{goodfellow2014generative,kingma2013auto,rudolph2019structuring,lecun1989generalization} are statistical models that learn the data distribution in an unsupervised manner. These models are able to generate samples from the training data manifold. Thus, anomaly detection occurs because anomalies can not be generated since they do not belong to the training set. Example of these generative models are an autoencoder\cite{kingma2013auto} and GAN \cite{goodfellow2014generative}.
Autoencoder-based approaches encode the high dimension features to their  latent dimension and reconstruct it back at the decoder end. Then, reconstruction error is calculated between the input and the output of autoencoder. Thus, a high reconstruction error should reflect there is an anomaly. Bergmann \andothers~\cite{bergmann2019mvtec} proposes to use SSIM (structure similarity index) as a loss to train the autoencoder. Thus, the SSIM index is used as a reconstruction error to capture visual similarities among the training data.  The main problem of the autoencoder-based approach is that they can generalize too strongly; as a result, they can reconstruct the anomalies as good as for the normal samples. Therefore, Gong \andothers~\cite{chen2005simultaneous} proposed tackling the generalization problem using a memory module to discretise the latent space. Zhai \andothers~\cite{zhai2016deep} propose to use regularized autoencoder with an energy-based model for modeling the data distribution of normal samples. Samples with high energy are considered as an anomaly.

GAN-based methods detect anomalies based on the assumption that GAN model can only generate normal samples.  GAN models consist of two models, namely generator and discriminator. Schlege \andothers~\cite{schlegl2019f} propose two-stage training approach for anomaly detection. Firstly, the GAN is trained, then  an encoder is optimized as an inverse generator. The idea of using the generator as a decoder enables the calculation of reconstruction error. The reconstruction error with the difference in discriminator features of the input and reconstructed image are both used as anomaly score. Akcay \andothers~\cite{akcay2018ganomaly} exploits the advantage of adversarial training by making autoencoder act as a generator of the GAN. The main goal here is to force the autoencoder to generate only the normal samples, which can be done by minimizing the reconstruction error between the embedding and original of the reconstructed data.

\subsubsection{Defect Detection with Pre-trained Networks}
These methods use the feature space of a pre-trained network to detect anomalies. Pre-trained networks use deep representation learning, particularly convolution neural networks that have been trained to discover multiple levels of representation in a supervised approach, such as classification tasks. The feature space of these networks is often generic enough that can be transferred to dissimilar task and still achieve competitive results \cite{donahuedeep}.  Anomaly detection using a pre-trained network is usually done by using a simple machine learning approach.  For example, Andrews \andothers~\cite{andrews2016transfer} use a simple One-Class-SVM (OCSVM) on VGG~\cite{simonyan2014very} features of the training images. Nazare \andothers~\cite{nazare2018pre} evaluated different pre-trained networks with different normalization techniques and perform anomaly detection by applying 1-Nearest-Neighbor classifier on the reduced PCA features. Sabokrou \andothers~\cite{sabokrou2018deep} model the normal features extracted from a pre-trained network using unimodal distribution. Rudolph \andothers~\cite{rudolph2021same} propose to model the distribution of pre-trained features using normalized flow network in which anomaly detection is performed based on the likelihood, lower likelihood as higher anomaly score.
\subsection{Point-pattern-based feature extraction}
\begin{figure*}
	
	\includegraphics[width=\textwidth]{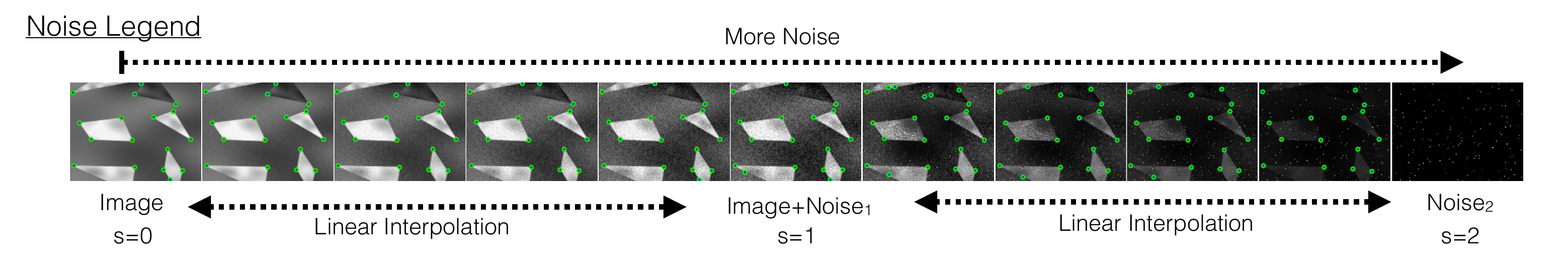}
	\caption{Superpoint point pattern feature extraction robustness against noise~\cite{detone2017toward}.}
	\label{Fig:sp_noise}

\end{figure*}

Point-pattern, commonly known as Keypoints, detection is regarded as a local feature extraction method. Local features extraction methods have many applications ranging from image retrieval~\cite{teichmann2019detect} , 3D reconstruction~\cite{schonberger2016structure} , camera pose estimation~\cite{sattler2019understanding} and medical image application~\cite{busam2018markerless}. These show the advantage of using sparse features over a direct network (tensor-based methods). Local feature extraction can be classified into handcrafted and learned-based methods.
\paragraph{Handcrafted Detectors}
Classical local feature detection or keypoints detection perform keypoints detection and descriptor computation independently. The localization of these keypoints is done by an engineered algorithm by looking at the geometric structure in the images. SIFT~\cite{lowe2004distinctive} extracts keypoints by finding blobs over multiscale levels on images and uses the gradient of the histogram as a descriptor. Harris~\cite{harris1988combined} and Hessian~\cite{beaudet1978rotationally} detector use the first and second-order derivatives to locate corners or blobs in images. Later, a multi-scale and affine transformations version of these detectors has been proposed in~\cite{mikolajczyk2004scale} and~\cite{tuytelaars2008local}. An acceleration of detection process via using the integral images as an approximation of the Hessian matrix has been proposed~\cite{bay2008speeded} which is known as SURF. A-KAZE~\cite{alcantarilla2011fast} keypoint detection uses a Hessian detector applied to non-linear diffusion scale space in contrast to commonly used Gaussian pyramid. Finally, MSER~\cite{matas2004robust} detected keypoints by segmenting the image and looking for stable regions.
\paragraph{Representation learning Detectors}
These methods are inspired by the success of representation learning in general object detection methods and feature descriptors. The earliest attempt to use machine learning for corner keypoints detection was made by FAST~\cite{rosten2006machine}. Different works have been proposed earlier to extend FAST by either optimizing it \cite{leutenegger2011brisk}, or adding a descriptor and adding orientation estimation~\cite{rublee2011orb}. The recent advances in convolutional neural networks (CNN) for representation learning have also made its impact on keypoints detection. TILDE~\cite{verdie2015tilde} uses CNN to detect keypoints that are robust under severe weather and illumination changes trained on multiple piece-wise linear regression models. Another approach to train CNN for keypoints detection is by using covariant constraints~\cite{lenc2016learning}.  Another attempt to make the training of CNN more stable by adding predefined detector anchors presented in~\cite{zhang2017learning}. Barroso-Laguna \andothers~\cite{barroso2019key} propose a KeyNet that uses handcrafted filters and learned filters together to detect keypoints.  The above approaches mainly focus on keypoints only. DeTone \andothers~\cite{detone2017toward} proposed two deep networks named MagicPoint and MagicWarp. MagicPoint extracts salient points and MagicWarp parameterise a transformation between two pair of images. Later, a self-supervised deep joint point detector and descriptor was proposed~\cite{detone2018superpoint} named as SuperPoint. SuperPoint network is trained on synthetic shapes (rendered lines, cubes, stars, triangles, quadrilaterals, checkboards) to detect corners. SuperPoint is the best option if you are interested in detecting corners in very noisy images, as shown in Fig. ~\ref{Fig:sp_noise}, and also feature detection and description are done in a self-supervised manner. The disadvantage of this approach is that it accept only a grayscale image. Thus, a colour-based defect cannot be detected. LIFT~\cite{yi2016lift} proposed end-to-end learning of keypoints detection and description including orientation estimation of every feature. LF-net~\cite{ono2018lf} estimates the scale, orientation, and position of features by joint learning of detector and descriptor. The LF-net trains the then network by pair of images (one of them is went through a homographic transformation) with a non-differential backpropagation for detection and the descriptor trained by triplet loss. Instead of using local maxima in detection keypoints, R2D2~\cite{revaud2019r2d2} train the network to detect keypoints that are reliable and repeatable and have uniform coverage of the image. R2D2 uses dilated convolution to preserve the spatial resolution but with very GPU and memory usage. The above approaches use the detect-then-describe. In which, the keypoints detection network is responsible for detecting low-level features such as corners or blobs under different scales, rotations and viewpoint changes. While, the descriptor network is responsible on high-level information by extracting patches around these keypoints. The problem of such approach is the luck of repeatability of these keypoints under strong appearance changes due to the fact the local detector considers small image region and can significantly get affected by small changes in pixel intensities.  Another possible approach is to use what is known as describe-then-detect approach~\cite{tian2020d2d}.  D2-Net~\cite{dusmanu2019d2} uses a single CNN network for joint detection and description using describe-to-detect approach.  The detection based on the local maxima across the channels and spatial feature maps

\section{Background}
\label{Sec:background}
This section outlines the background theory of energy-based models and random finite set theory.
\subsection{Energy-based models}
Energy-based models (EBMs) are a family of statistical models that use energy function $E(\mathbf{x};\theta)$ to represent probability distribution through unnormalized negative log probability. The density function for input $\mathbf{x}\in R^d$ in the form:
\begin{equation}
p(\mathbf{x};\theta)=\frac{e^{-E(\mathbf{x};\theta)}}{Z(\theta)},
\end{equation}
where $Z(\theta)=\int_{\mathbf{x}}e^{-E(\mathbf{x};\theta)}d\mathbf{x}$ is the normalized function to ensure the density function $\int p(\mathbf{x})d\mathbf{x}=1$; $\theta$ are the model parameters to be learned. The nice property of EBM model is that one is free to choose any suitable parametrization of the energy function. Parameters learning of EBM models can be done by assigning lower energy (hence high probability) to the observed samples. Learning via Maximum likelihood estimator (MLE) is impossible due to the intractability of partition of $Z(\theta)$. Thus, MCMC approximation methods of the integral are applied by the summation over samples from a Markov chain~\cite{zhai2016deep}. 
\subsection{Random finite set-based anomaly detection}

Random finite sets statistics is one of the stochastic geometric models, which is a well-established study area that dates back to famous Buffon's needle problem~\cite{chiu2013stochastic}. Random Finite Sets (RFSs) are set-valued random variables with an unknown number of elements that are themselves random. Within the scope of this paper, there are a set of keypoints and their corresponding descriptors for each particular image. Given an underlying space $\mathcal{X}$ as keypoints space, a random finite set draw instantiation from the hyperspace of all finite subsets $\mathcal{F(X)}$ of $\mathcal{X}$. Generally, a random finite set may contain a finite number of elements, possible realizations of a random finite set $X \in \mathcal{F(X)}$ are

\begin{equation}
X=\emptyset,X= \left\lbrace \mathbf{x}_1 \right\rbrace ,X=\left\lbrace \mathbf{x}_1,\mathbf{x}_2 \right\rbrace ,\cdots X=\left\lbrace \mathbf{x}_1,\cdots, \mathbf{x}_n \right\rbrace , 
\end{equation}
where $\mathbf{x}_i \in \mathcal{X}$ and $\forall_i,j:\mathbf{x}_i\neq \mathbf{x}_j$. Also, RFS impose no ordering on their set elements. In RFS-based visual anomaly detection, the measurements are modeled as an RFS. Hence, each extracted set of features $X=\{ \mathbf{x}_i\}_{i=1}^{|X|}$ is treated as an RFS, where $\mathbf{x}_i \in \mathbb{R}^D$ is a high dimension real continuous random variable. The density of RFS with respect to dominating measure $\nu$ is given by~\cite{vo2018model}:

\begin{equation}
p(X)=p(|X|)(|X|)!V^{|X|} p_{|X|}( \mathbf{x}_1,\cdots,\mathbf{x}_{|X|}),
\end{equation}
where $|X|$ is the cardinality of the set $X$, $p(|X|=n)=p(n)$ is the discrete cardinality distribution, $V$ is the unit hyper-volume, and $p_{n}( \mathbf{x}_1,\cdots,\mathbf{x}_{n})$ is the symmetric joint feature density for given cardinality $|X|=n$.

In practice, difference assumptions can be considered regarding the mathematical form of the RFS density $p(X)$. For instance, Poisson RFS~\cite{1261119Mahler}, Beta RFS~\cite{Kamoona9074564}, the Bernoulli RFS~\cite{mahler2007statistical}, the multi-Bernoulli RFS~\cite{vo2008cardinality}, and finally the generalized labeled mutli-Bernoulli RFS~\cite{papi2015generalized}. Most of densities, mentioned earlier, are used to model mutli-object entity as an RFS. Another simplified assumption about RFS form is an independent identical distributed (IID) cluster RFS density as follows~\cite{Vo2018,kamoona2019random}:

\begin{equation}
p(X)=p(|X|)(|X|)!V^{|X|}[p(\cdot)]^{|X|},
\end{equation}

where $p(\cdot)$ is single feature density and $[p(\cdot)]^{|X|} \triangleq \prod _{x\in X} p(\mathbf{x})$ is a finite set exponential. Given the discrete cardinallity distribution $p(|X|)$ follows the Poisson distribution, the IID-cluster RFS turn into Poisson RFS given by:
\begin{equation}
p(X)=\rho^{|X|} \exp{(-\rho)}\left[U\right] ^X \prod _{x\in X} p(\mathbf{x}),
\label{eq:poisson}
\end{equation}
where $\rho >0$ is Poisson intensity.
\subsubsection{Parameters learning of IID-RFS cluster density}
\label{sec:IDD_RFS_learning}
Parameters learning of IID-RFS cluster density is similar to parameter learning of probabilistic generative models~\cite{goodfellow2016deep} which is done by maximizing the log-likelihood of normal samples. Given IID cluster RFS, the goal is to estimate the following parameters:

\begin{equation}
\begin{split}
(\hat{\rho},\hat{\theta})~& = \argmax_{\rho,\theta} \prod_{X\in {\bigO}}\mathcal{L}(\rho,\theta|X), \\
& = \argmax_{\rho,\theta} \prod_{X\in {\bigO}}\mathcal{L}(X|\rho,\theta)~p(\rho,\theta),
\end{split}
\label{eq:IDD_RFS_paramer}
\end{equation}
where $(\rho,\theta)$ are parameters of the cardinality as $p_c(|X|;\rho)$, and multi-feature joint density as $p(\mathbf{x}_1,\ldots,\mathbf{x}_{|X|};\theta)$ respectively,  $\mathcal{L}(X;\rho,\theta)$ is the likelihood function that measure the plausiblity of $(\rho,\theta)$, given some observed set $X$,where ${\bigO}$ is the ensemble of all training feature sets,  and finally $p(\rho,\theta)$ is the prior distribution which assumed uniform in most applications. In light of this, Eq.~\ref{eq:IDD_RFS_paramer} can be more simplified to the following:

\begin{equation}
\begin{split}
(\hat{\rho},\hat{\theta})~& = \argmax_{\rho,\theta} \prod_{X\in {\bigO}}\mathcal{L}(\rho,\theta|X).\\
\end{split}
\label{eq:IDD_RFS_paramer_2}
\end{equation}
Eq.~\ref{eq:IDD_RFS_paramer_2} can be re-written for Poisson RFS density as follows:
\begin{equation}
(\hat{\rho},\hat{\theta}) = \argmax_{\rho,\theta} \prod_{X\in\bigO} p(|X|;\rho) |X|! U^{|X|}\left[p(\cdot;\theta)\right]^{X}.
\end{equation}

Vo~\andothers~\cite{vo2018model} have proven that the aforementioned optimization can be turned into two different optimizations one for cardinality and one for feature density as:
\begin{eqnarray}
\hat{\rho} & = & \argmax_{\rho} \prod_{X\in\bigO} p(|X|;\rho) \label{eq:theta_c1} \\
\hat{\theta} & = & \argmax_{\theta} \prod_{X\in\bigO} \left[p(\cdot;\theta)\right]^{X}. \label{eq:theta}
\end{eqnarray}
Substituting $p(|X|;\rho)$ with $\rho^{|X|}\,e^{-\rho}\big/\,|X|!$ in Eq.~\ref{eq:theta_c1}, the solution simply turns out as average cardinality of all training feature sets values as follows:
\begin{equation}
\hat{\rho} = {\sum_{X\in\bigO} |X|}\bigg/{|\bigO|}.
\label{eq:rhohat}
\end{equation}
Assuming that all point features within each feature set are IID vectors in $\mathbb{R}^D$ and distributed according to a Gaussian density with parameters ${\theta} = (\mu,\Sigma)$, Eq.~\eqref{eq:theta} turns into:
\begin{eqnarray}
(\hat{\mu},\hat{\Sigma}) = \argmax_{(\mu,\Sigma)} \prod_{X\in\bigO} \left[\prod_{\mathbf{x}\in X}\mathcal{N}(\mathbf{x};\mu,\Sigma)\right],
\end{eqnarray}
where $\mathcal{N}(\cdot;\mu,\Sigma)$ is the multivariate Gaussian density function with mean $\mu \in \mathbb{R}^D$ and covariance $\Sigma\in \mathbb{R}^{D\times D}$. The above optimization has a closed-form solution as follows:
\begin{eqnarray}
\hat{\mu}&=& {\sum_{X\in\bigO}\sum_{\mathbf{x}\in X}\mathbf{x}}\ \bigg/\ {\sum_{X\in\bigO}|X|}\label{eq:muhat}\\
\hat{\Sigma}&=& {\sum_{X\in\bigO}\sum_{\mathbf{x}\in X}(\mathit{\mathbf{x}}-\hat{\mu})(\mathit{\mathbf{x}}-\hat{\mu})^\top}\bigg/{\sum_{X\in\bigO}|X|}.\label{eq:Sigmahat}
\end{eqnarray}
Eqs. \eqref{eq:muhat} and \eqref{eq:Sigmahat} are used to estimate the mean and covariance of a single Gaussian component.

From Eq.~\ref{eq:muhat} and~\ref{eq:Sigmahat}, it is obvious that the estimated mean $\hat{\mu}$ and covariance $\hat{\Sigma}$ for single feature density are simply the sample mean and covariance respectively. 
To avoid the singularity problem of the covariance estimate on the high dimension space  $\hat{\Sigma}\in \mathbb{R}^{D\times D}$, Vo~\andothers~\cite{vo2018model} use PCA algorithm to reduce the dimension of the training features as preprocessing step. However, the problem of this approach is that it suffers from decoupled model learning and incapability of preserving essential information~\cite{zong2018deep} due to two-step approaches, dimension reduction and parameters learning. We avoid this, by performing parameters learning in the high dimension space as shown in section~\ref{Sec:RFS_energy_parameter_learning} which does not require any training phase and can be estimated given few training samples as proven in few-shot learning experiments, see section~\ref{Sec:Fewshot_results}.

\section{Proposed Approach}
\label{Sec:proposed_appraoch}
\begin{figure*}
	\includegraphics[width=\textwidth]{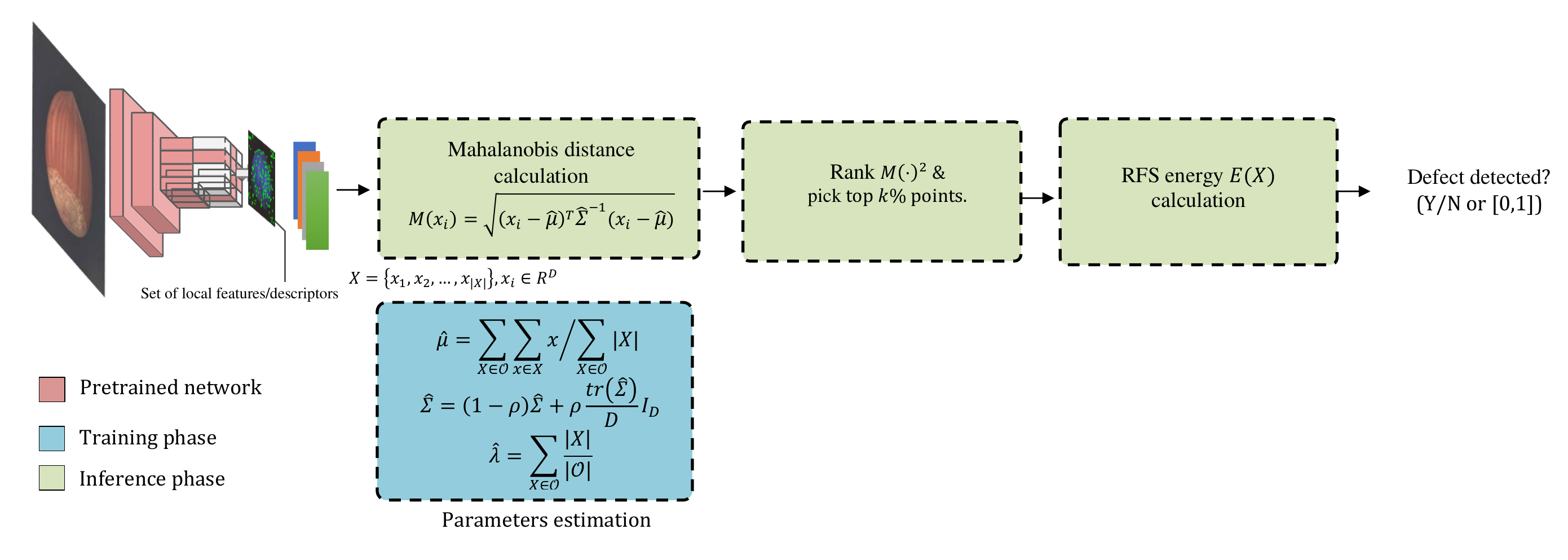}
	\caption{Our proposed energy-based RFS of point pattern features for anomaly detection.}
	\label{fig:proposed_approach}
\end{figure*}

We propose to use energy-based RFS for detect detection, illustrated in~Fig.~\ref{fig:proposed_approach}. The proposed approach mainly consists of two parts: local features extraction backbone and RFS energy calculations. The proposed approach is very computationally efficient and does not require a heavy training computation. For feature extraction, we use a pre-trained CNN local features network to capture the geometrical information of the object. We choose to model a set of point pattern features as Random Finite Set (RFS), the RFS energy is computed for image point pattern features.

\subsection{Feature extraction backbone}

In anomaly detection, we are interested in detecting the irregularities in the image, and these variations may be very small. One approach to capture these irregularities is the use of global features. However, the problem of these networks can be affected by a wide range of conditions, such as lighting and viewpoint changes. Thus, we propose to use a local feature extraction pipeline, namely D2-Net~\cite{dusmanu2019d2}. This network uses the describe-and-detect approach.

Given an input image $I$, a D2-Net feature extractor network $\mathcal{F}$ produce a 3D tensor $F=\mathcal{F}(I)$,  $F \in\mathbb{R}^{h\times w\times n}$, where $h\times w$ is the spatial dimension of the feature map and $n$ is the number of the channels. The 3D tensor $F$ can be seen as set of collection of 2D responses $D$:

\begin{equation}
D^k=F_{::k},D^k \in \mathbb{R}^{h\times w\times n},
\end{equation}
where $k=1,\ldots,n$. We can see that the feature extraction $\mathcal{F}$ network generates $n$ different response maps $D^k$. These response maps are similar to its analogue counter apart  Difference-of-Gaussian response maps in SIFT. To detect a location of keypoint and their descriptor, D2-Net defines detection at point $(i,j)$ if  it satisfies the following:

\begin{equation}
\begin{split}
(i,j)~\text{is a detection} \Leftrightarrow D^k_{i,j} \text{~is a local max in} D^k,\\
\text{with~} k=\argmax_{t}  D^t_{i,j}
\end{split}
\end{equation}

\subsection{Point pattern RFS energy}
The backbone pretrained features extraction network returns a set of local point pattern features, and we choose to model these features as a RFS. For RFS anomaly detection, the general approach for modeling the likelihood of new measurement $X^*$ is via RFS log-likelihood as follows:
\begin{equation}
\mathcal{L}(X^*;\rho,\theta)=\log(p(|X^*|;\rho))+\log(|X^*|!)+\log\prod_{\mathbf{x}\in X^*} p(\mathbf{x};\theta)
\label{RFS+Pos_log_like}
\end{equation}
where $\mathcal{L}(X^*;\rho,\theta)$ is the RFS log-likelihood, $\rho,\theta$ are the cardinalilty and joint feature density parameters respectively. With Poisson assumption for cardinalilty and multivariate Gaussian distribution, $\mathcal{N}(\mathbf{x};\mu,\Sigma)$ for single feature density, the RFS log-likelihood turns into:

\begin{equation}
\mathcal{L}(X^*;\rho,\mu,\Sigma)=\log(\frac{\rho\exp(-\rho|X^*|)}{|X^*|!})+\log(|X^*|!) \\
+\log\prod_{\mathbf{x}\in X^*} \mathcal{N}(\mathbf{x};\mu,\Sigma).
\end{equation}
Due it unit inconsistency, the above log-likelihood can not be used for ranking the point pattern features. To overcome this, Vo~\andothers~\cite{Vo2018} proposed RFS ranking $r(X)$ as follows:
\begin{equation}
\label{eq:rankrfs}
r(X)= p(|X|) \left[\frac{p(\cdot)}{||p(\cdot)||_2^2}\right] ^X,
\end{equation}
where $||p(\cdot)||_2^2$ is the squared $L^2$-norm of $p(\cdot)$. The above ranking function is based on the assumption that ranking function $r(X)~\propto p(|X|)~p(X)$. Calculating the $L_2$ in high dimension is not a feasible solution due to the singularity problem. Also, the RFS ranking function can not capture the small variation caused by defects due to the $L_2$ normalization.

Inspired by the current success of applying the energy model for anomaly detection~\cite{zhai2016deep,grathwohl2019your,liu2020energy}, we propose the use of RFS energy as an efficient approach for defect detection.
The proposed RFS energy function for IID RFS is based on the assumption that RFS energy~$E(X) \propto E(p(|X|))\times E(p(X))$ is proportional to both cardinality energy and single feature energy as follows:
\begin{equation}
E(X;\rho,\mu,\Sigma)=-\log(\rho^{|X|})+\log(|X|!) +\sum_{\mathbf{x}\in X} M(\mathbf{x};\mu,\Sigma)^2,
\label{RFS_engery}
\end{equation}
where $E(\cdot)$ is the Poisson RFS energy, $M(\cdot)^2$ is the squared Mahalanobis distance:
\begin{equation}
M(\mathbf{x};\mu,\Sigma)^2=(\mathbf{x}-\mu)\Sigma^{-1}(\mathbf{x}-\mu)^T.
\label{Mah_dis}
\end{equation}
The Mahalanobis distance is a point to distribution distance introduced in 1936~\cite{mahalanobis1936generalized} which is very well-known for modeling the sample's uncertainty. Recent works show the effectiveness of using Mahalanobis distance to model the pre-trained features of normal samples \cite{christiansen2016deepanomaly,rippel2021modeling}. However, to the best of our knowledge, no work has used the sum squared of Mahalanobis distance to model local point pattern feature of normal samples within RFS framework.

\subsection{Parameters learning of RFS energy}
\label{Sec:RFS_energy_parameter_learning}
To learn the parameter of RFS energy without countering the problem of decoupled learning~\cite{zong2018deep}, we choose to learn the parameters on the high dimension space without further processing, such as PCA feature reduction. In addition, we are also inspired by the current success~\cite{lee2018simple,rippel2021modeling} of using pre-trained global features for anomaly detection without fine-tuning. For IID cluster Poisson RFS energy (Eq.~\ref{RFS_engery}), we have three parameters to learn, which are the Poisson intensity $\rho$, mean $\mu$, and covariance $\Sigma$ of multivariate Gaussian density. We follow the same approach mention in section ~\ref{sec:IDD_RFS_learning} to learn the parameters by considering the parameter learning of cardinality and feature density are two separate optimization problems; see equations~Eq. (\ref{eq:rhohat},~\ref{eq:muhat}~ and\ref{eq:Sigmahat}).

The main significant difference here is that parameter learning of the feature density is done on the high dimension space $\mu\in \mathbb{R}^D.\Sigma \in \mathbb{R}^{D\times D}$. Since the true distribution of features is unknown, the mean and covariance need to be leaned. However, covariance estimation using Eq. ~\ref{eq:Sigmahat} requires the number of  training features samples $\mathbf{x},\ldots,\mathbf{x}_N$, say $N$, to be much larger from feature dimension $N>>D$. Accordingly, when there are few normal samples, which is the case of most anomaly detection settings and particularly in few-shot learning, the estimation becomes unstable and leads to the singularity problem. To overcome this problem, we use shrinkage covariance estimation~\cite{ledoit2004well} as follows:

\begin{equation}
\hat{\Sigma}_{shrunk}=(1-\alpha)\hat{\Sigma}+\alpha\frac{tr(\hat{\Sigma})}{D} I,
\label{eq:Sigma_shrunk}
\end{equation}
where $\hat{\Sigma}$ is the empirical estimated covariance, $\alpha$ is the shrinkage intensity, and $I$ is the identity matrix. Shrinkage estimation is a linear combination of empirical estimated covariance $\hat{\Sigma}$ and the scaled identity matrix $I$ and $\alpha$ regulates this influence on the final matrix. Ledoit~\andothers~\cite{ledoit2004well} derived a closed-form solution by minimizing the expected squared error $E[||\hat{\Sigma}_{shrunk}-\Sigma||^2]$ for the amount of the shrinkage allowed to obtain the optimal parameter of $\alpha$ given the unstable covariance estimate $\hat{\Sigma}$.

\subsection{Choosing the $top~k$ squared Mahalanobis distances}
In the defect detection problem, we are looking for small irregularities in RFS energy caused by a defect in a small region in the image. Due to the sum operation in RFS energy, capturing the small variation is close to be negligible. The small variation is reflected via the Mahalanobis distance from the normal feature distribution. the point pattern descriptor of the defect region definitely has a greater Mahalanobis distances compared to normal one. For this reason, we propose to use the features that have higher distance from the normal distribution in our calculation. We implemented this by ranking the squared distance Mahalanobis and choose only the $top~k$ percent in the RFS energy calculation.

\section{Experiments}
\label{Sec:experiments}
\subsection{Dataset}
\begin{figure}[ht!]
	\includegraphics[width=0.47\textwidth]{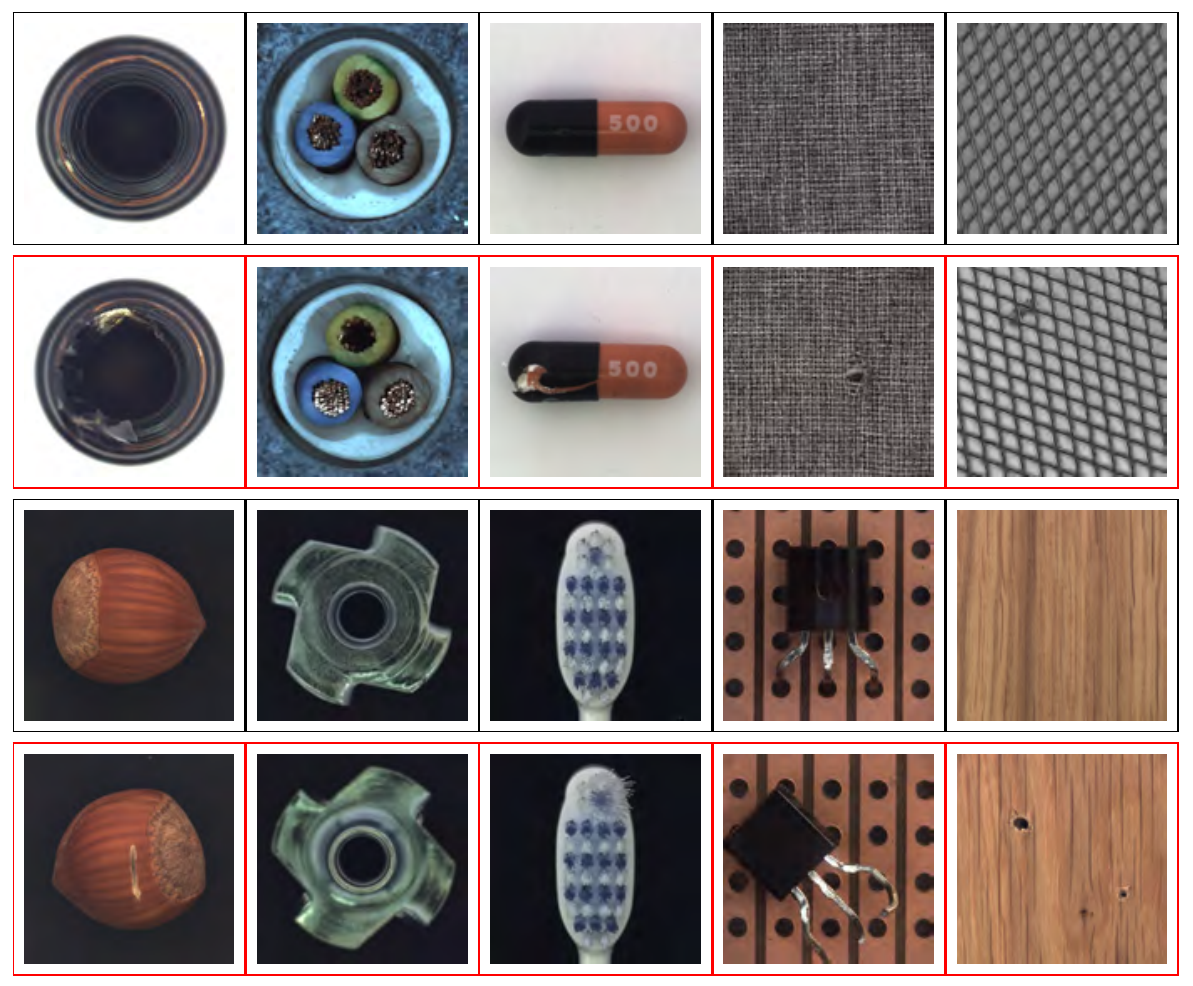}
	\caption{Samples of MVTec-AD dataset, defect free (black box) and defected samples (red box).}
	\label{Mvtec}
\end{figure}

\textbf{MVTec-AD dataset}:
MVTec-AD~\cite{bergmann2019mvtec} is a real-world industrial image dataset that has been developed as a comprehensive and challenging benchmark for defect detection. The dataset has a large-scale collection of texture and object images. In general, it contains 5354 high-resolution color images of 15 different categories (ten objects and five textures). Each object has normal samples and defect samples. The training set contains only normal samples, and the test set contains both normal and defect samples. Seventy different types of defects differ in size, shape and structure, including scratches, dents, contamination, and various structural deformations present in this dataset.
Fig.~\ref{Mvtec} shows a set of these samples. The first row shows the defect-free samples, while the second row shows defect samples.

\subsection{Implementation Details}

In all experiments, we use pre-trained CNN-based for point pattern feature extraction. We have used a D2-Net network for feature extraction, and we have examined different other networks, more details in the ablation study. Before passing the image to the D2-net, we have resized all images into 256$\times$256 and cropped at the center which yields in image size $224 \times 224$. In addition, we use the multi-scale option of D2-net in the feature extraction phase, which scale the image into three scales (0.5, 1, and 2). Our model does not require any training phase, and parameter learning of the multivariate RFS energy has closed-form for the mean of RFS features and cardinality of the features set. The only computation performed is in estimating the covariance of the feature density using shrinkage method ~\cite{ledoit2004well}.
\subsection{Evaluation protocol}
We follow the same protocol defined in ~\cite{rudolph2021same,ganomaly,rippel2021modeling} to fully assess the capability of the proposed approach by reporting the Area Under Receiver Operator Characteristic (AUC). The AUC measures the area under the true positive rate as a function of the false positive rate. The advantage of using AUC is that not sensitive to any threshold or percentage of anomalies present in the test set.
\subsection{Experimental results}
\begin{figure}[!ht]
    \centering
    \includegraphics[scale=0.5]{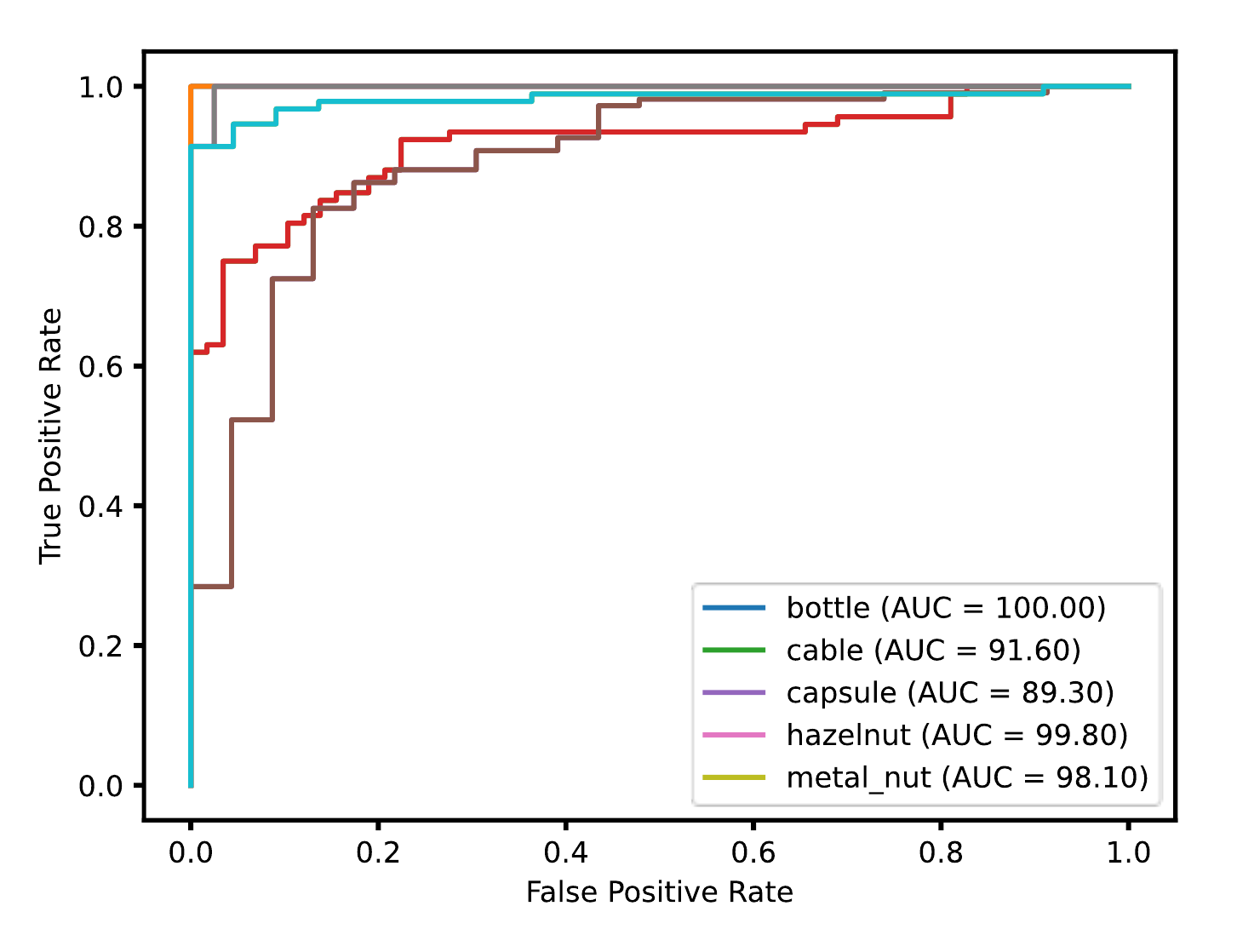}
    \caption{ROC curve for bottle, cable, capsule, hazelnut, and metalnut using D2-Net features.}
    \label{fig:ROC_curve_mvtec_1}
\end{figure}

\begin{figure}[!ht]
\centering
    \includegraphics[scale=0.5]{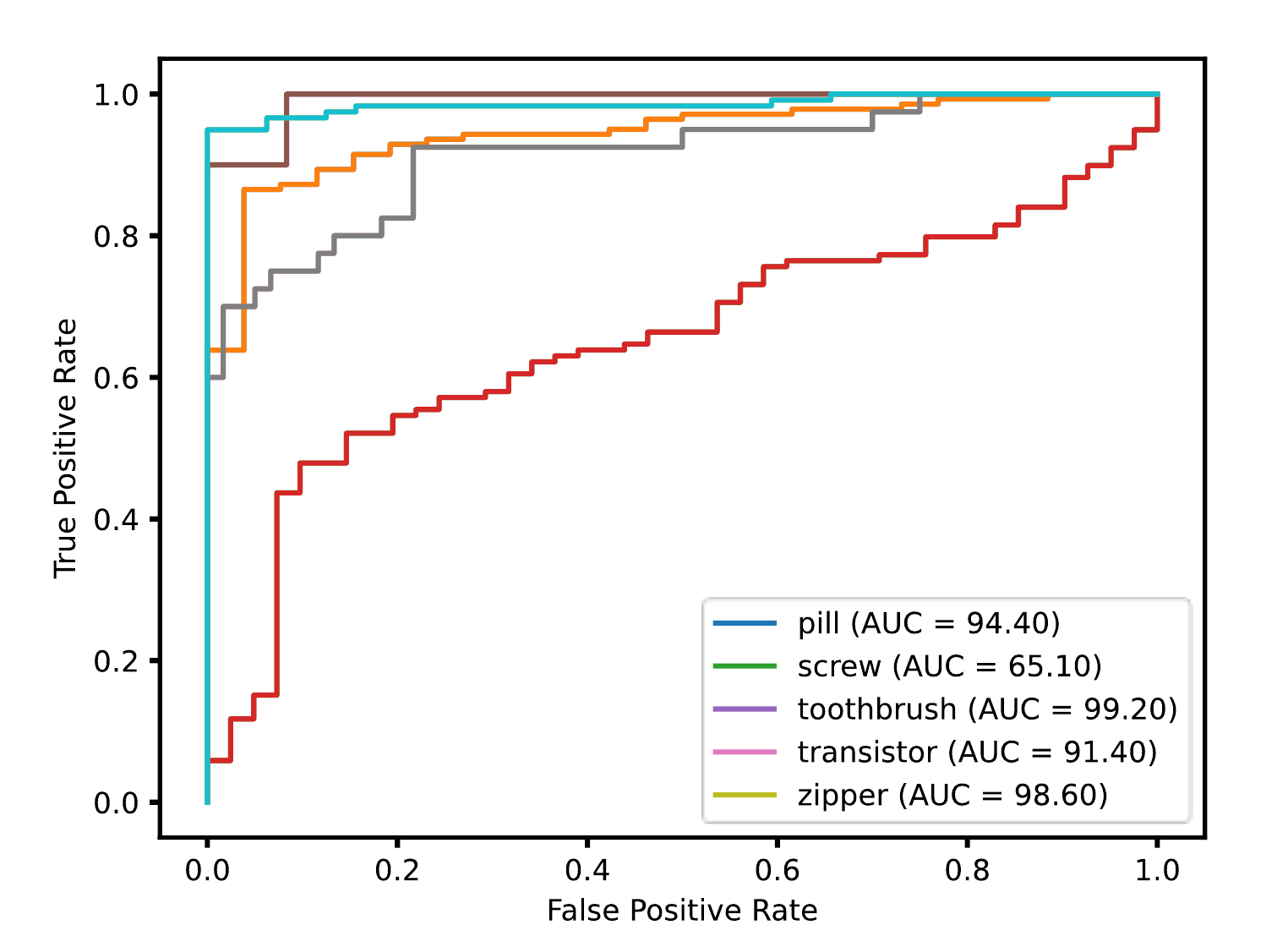}
      \caption{ROC curve for pill, screw, toothbrush, transistor, and zipper using D2-Net features.}
    \label{fig:ROC_curve_mvtec_2}
\end{figure}

\begin{figure}[!ht]
\centering
     \includegraphics[scale=0.5]{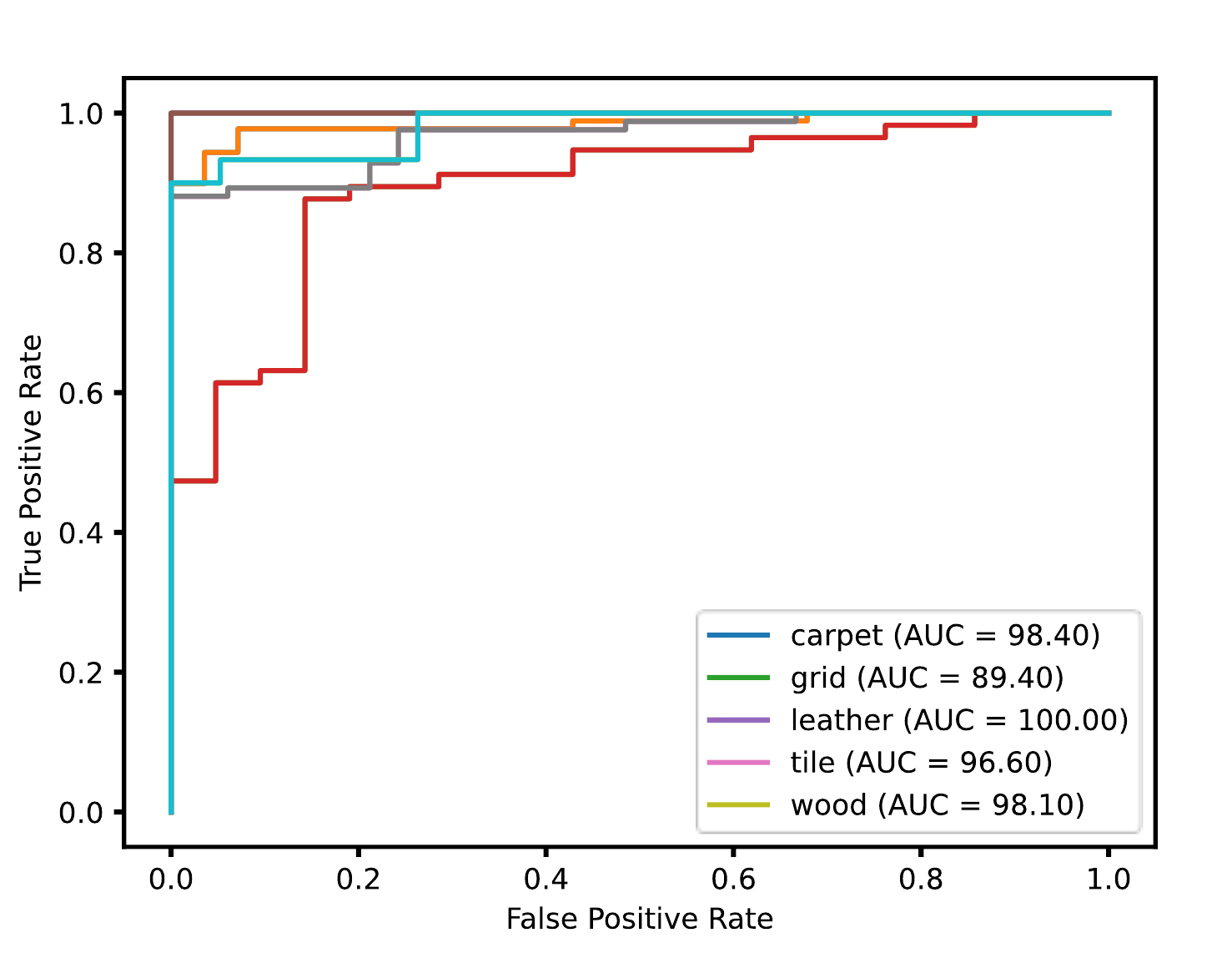}
    \caption{ROC curve for carpet, grid, leather, tile, and wood using D2-Net features.}
    \label{fig:ROC_curve_mvtec_3}
\end{figure}

In this section, we explore the performance of the proposed approach against different deep learning models that use global features for defect detection. We evaluate the proposed approach on the MVTec AD dataset and report the AUC accuracy. We compare the proposed approach with two transfer learning-based models, One-Class SVM (OCSVM)~\cite{andrews} that used transfer learning of CNN global feature, the use of distance to the nearest neighbour (1-NN) of the CNN features, and the z-score normalization~\cite{nazare}. Also, we compare our model with non-transfer learning methods such as GeoTrans~\cite{geotrans}. GeoTrans calculate the anomaly score based on the classification of conducted geometrical transformations that alleviate the need for a generative model. GANomaly~\cite{akcay2018ganomaly} that use adversarial learning (GAN) by exploiting the autoencoder as the generator of the GAN, which forces the decoder to generate only normal samples. DSEBM~\cite{dsebm} that uses the energy model for anomaly detection of detect. Also, we compare our approach with the recent normalized flow approach known by DifferNet~\cite{rudolph2021same}. Table ~\ref{Tab:MVTec_AUC} shows the results for MVTec AD dataset. Despite the fact that the proposed RFS energy-based does not require any training, the proposed approach outperforms most existing models that use global features. There is only a very low margin 0.2 compared to current state-of-art DifferNet~\cite{rudolph2021same}. However, our approach outperforms all methods for 11 out of 15 objects and achieves the second-best performance for the rest, 3 out of 4. The lowest performance of the proposed approach lies in the screw object. It achieves only (AUC=70), and we argue the main reason for this is due to it the feature extraction backbone. Better performance has been observed for this object using other point pattern feature extraction backbone, see the ablation study. The ROC curves of our proposed approach for 15 categories are shown in Figs.~\ref{fig:ROC_curve_mvtec_1},\ref{fig:ROC_curve_mvtec_2}, and \ref{fig:ROC_curve_mvtec_3}. We can observe a high true positive rate of our proposed approach for most objects. The source code of the paper is available at\footnote{ \url{https://github.com/AmmarKamoona/RFS-Energy-Anomaly-Detection-of-Defect}}

\begin{table*}[!h]
	\caption{AUC in \% for detected anomalies of all categories of MVTec AD grouped into textures and objects. Best results are in bold, second best underlined.}
	\label{Tab:MVTec_AUC}
	\centering
\begin{tabular}{@{}c@{}c@{}} 
	\begin{tabular}{ccccccccc}
		    \toprule
		& Category & GeoTrans & GANomaly & DSEBM & OCSVM & 1-NN & DifferNet & \textbf{RFS energy}\\
		& & \cite{geotrans} &\cite{akcay2018ganomaly}  & \cite{dsebm} & \cite{andrews}& \cite{nazare}&\cite{rudolph2021same} &(our)\\
		\cline{2-9}
		
		& Bottle     & 74.4 & 89.2 & 81.8 & 99.0& 98.7 & 99.0  &\textbf{100}\\
		& Cable      & 78.3 & 75.7 & 68.5 & 80.3 & 88.5 & \textbf{95.9} &\underline{92.0}\\
		& Capsule    & 67.0 & 73.2 & 59.4 & 54.4 & 71.1 & 86.9 &\textbf{89.4}\\
		& Hazelnut   & 35.9 & 78.5 & 76.2 & 91.1 & 97.9 & 99.3 &\textbf{99.9}\\
		& Metal Nut  & 81.3 & 70.0 & 67.9 & 61.1 & 76.7 & 96.1 &\textbf{98.2}\\
		\rotatebox[origin=c]{90}{\parbox[c]{0cm}{Objects}}& Pill       & 63.0 & 74.3 & 80.6 & 72.9 & 83.7 & 88.8 &\textbf{94.5}\\
		& Screw      & 50.0 & 74.6 & \textbf{99.9} & 74.7 & 67.0 & \underline{96.3} &70.0\\
		& Toothbrush & 97.2 & 65.3 & 78.1 & 61.9 & 91.9 & 98.6 &\textbf{99.2}\\
		& Transistor &86.9 & 79.2 & 74.1 & 56.7 & 75.6 & 91.1 &\textbf{91.9}\\
		& Zipper     & 82.0 & 74.5 & 58.4 & 51.7 & 88.6 & 95.1 &\textbf{98.7}\\
		\cline{2-9}
		& Carpet     & 43.7 & 69.9 & 41.3 & 62.7 & 81.1 & 92.9 &\textbf{98.4}\\
		& Grid       & 61.9 & 70.8 & 71.7 & 41.0 & 55.7 & 84.0 &\textbf{89.64}\\
		& Leather    & 84.1 & 84.2 & 41.6 & 88.0 & 90.3 & 97.1 &\textbf{100}\\
		& Tile       & 41.7 & 79.4 & 69.0 & 87.6 & 96.9 & \textbf{99.4} &\underline{96.9}\\
		
		\rotatebox[origin=c]{90}{\parbox[c]{0cm}{Textures}}& Wood       & 61.1 & 83.4 & 95.2 & 95.3 & 93.4 & \textbf{99.8} &\underline{98.1}\\
		\cline{2-9}
		& \textbf{Average} & 67.2 & 76.2 & 70.9 & 71.9 & 83.9 &\textbf{94.7} &\underline{94.5}\\
		& \textbf{Median} &  &  &  &  &  &96.1 &\textbf{98.1}\\
		    \toprule
		
	\end{tabular}
\end{tabular}
\end{table*}	

\subsection{Study Analysis}
In this section, we further investigate the performance of our RFS energy compared to the following: the use of RFS log-likelihood as anomaly score~\cite{Vo2018} as shown in Eq. \eqref{RFS+Pos_log_like}, and the use of the sum of Mahalanobis distances only as anomaly score as follows:
\begin{equation}
    AS=\sum_{\mathbf{x}\in X} M(\mathbf{x};\mu,\Sigma),
\end{equation}
where $AS$ is the anomaly score of a set of the point pattern features. We compare the aforementioned anomaly scores with the proposed RFS energy, and we use the D2-Net features and the same learned parameters, $\mu \in \mathbb{R}^D,\Sigma \in \mathbb{R}^{D\times D}$. We report the AUC for MVTec AD dataset as shown in Table~\ref{Tab:Mal_Vs_rfs}. Table~\ref{Tab:Mal_Vs_rfs} shows the AUC of the proposed RFS energy approach achieves the best results in all categories compared to using $AS$ and RFS log-likelihood as an anomaly score, and this highlights the significance of using RFS energy as anomaly score.
\begin{table}[!h]
\centering
\caption{AUC in~\% for MVTec AD comparison between using RFS energy, RFS log-likelihood, and sum of mahalabonis distance $AS$ as anomaly scores.}
    \begin{tabular}{lccc}
    \toprule
    \textbf{Category} &RFS energy&$AS$&RFS log-likelihood \\
    \hline
    Bottle & \textbf{100.0} & 100.0 & 0.1 \\
    Cable & \textbf{92.0 } & 86.2  & 30.8 \\
    Capsule & \textbf{89.4}  & 81.4  & 36.7 \\
    Hazelnut & \textbf{99.9}  & 77.5  & 3.3 \\
    Metalnut & \textbf{98.2}  & 74.1  & 17.4 \\
    Pill  & \textbf{94.5}  & 85.9  & 48.8 \\
    Screw & \textbf{70.0 } & 4.6   & 0.6 \\
    Toothbrush & \textbf{99.2}  & 88.6  & 16.7 \\
    Transistor & \textbf{91.9}  & 80.8  & 9.3 \\
    Zipper & \textbf{98.7}  & 90.7  & 28.8 \\
     \hline
    Carpet & \textbf{98.4}  & 87.6  & 1.9 \\
    Grid  & \textbf{89.6}  & 58.7  & 40.5 \\
    Leather & \textbf{100.0} & 100.0 & 0.0 \\
    Tile  & \textbf{96.9}  & 89.5  & 8.0 \\
    Wood  & \textbf{98.1}  & 93.2  & 4.3 \\
    \hline
    Avg.  & \textbf{94.5}  & 79.9  & 16.5 \\
        \toprule
    \end{tabular}
    \label{Tab:Mal_Vs_rfs}
\end{table}

\subsection{Ablation Study}

In this section, we study the effect of different point pattern feature extraction methods on  the proposed RFS energy-based defect detection performance. All the deep point pattern features networks used in this experiment are pre-trained. The focus of this ablation study is on transfer learning of deep point pattern features. However, we also include one handcrafted feature extraction, namely an ORB\cite{rublee2011orb}.  General statistics of these methods/networks shown in Table~\ref{Tab:statistic_point_patern}. Table~\ref{Tab:statistic_point_patern} shows the name of the feature extraction method/network, the name of the datasets that have been used to train these models, the type of the input image, whether the network detects keypoints using color information or not, and the size of output descriptor of the network. Table~\ref{Tab:ablation_study} show results of different point pattern feature extraction methods using our RFS energy-based model. The results show that handcrafted ORB features have a very poor performance compared to other transfer learned-based features.
On the other hand, Superpoint features show a good performance but still does not achieve better performance than D2-Net features, and one reason because this network extract corner points and thus does not take into account the corner that results  due to color deformation. Another important point to notice is that R2D2 achieves the best performance for the screw object compared with others. Finally, we could achieve the best mean performance when we take the best performance features using our RFS energy-based model. 

\begin{table*}[!h]
\caption{General statistics of point pattern feature extraction methods used in the ablation study.}
\centering
\begin{tabular}{cccc}
	    \toprule
	Featute extraction &training dataset& input image& descriptor size\\
	\hline
	ORB&-& gray&256-D\\
	LF-Net~\cite{ono2018lf}~\footnote{\url{https://github.com/vcg-uvic/lf-net-release}}& ScanNet& gray& 256-D\\
	KeyNet~\footnote{\url{https://github.com/axelBarroso/Key.Net}}&synthetic training set from ImageNet \& 
	ILSVRC 2012 dataset&gray& 128-D\\
	Superpoint (SP)~\footnote{\url{https://github.com/rpautrat/SuperPoint}}&synthetic shapes \& COCO 2014& gray&256-D \\
	R2D2~\footnote{\url{https://github.com/naver/r2d2}}& Oxford and Paris retrieval dataset&color& 128-D\\
	    \toprule
\end{tabular}
\label{Tab:statistic_point_patern}
\end{table*}

\begin{table*}[!htbp]
	\centering
	\caption{AUC in \% for detected anomalies of all categories of MVTec AD using different point pattern feature extraction methods of our proposed RFS energy. The last two columns show the comparison between the best performance of our RFS energy and DifferNet.}
	\label{Tab:ablation_study}
	\begin{tabular}{ccccccccc}
		    \toprule
		Category &ORB &R2D2&LF-Net&KeyNet &SP &D2-Net&Best performance (Max) &DifferNet\cite{rudolph2021same} \\
		\hline
		Bottle & 63.9  & 50.0  & 93.7  & 94.9  & 99.4  & 100.0 & 100.0& 99.0  \\
		Cable & 44.1  & 42.0  & 82.8  & 63.0  & 83.2  & 92.0  & 92.0  & 95.9 \\
		Capsule & 47.3  & 70.4  & 58.7  & 66.7  & 74.0  & 89.4  & 89.4  & 86.9 \\
		Hazelnut & 83.2  & 79.9  & 86.0  & 85.3  & 94.9  & 99.9  & 99.9  & 99.3 \\
		Metalnut & 30.3  & 0.0   & 54.1  & 77.5  & 71.2  & 98.2  & 98.2  & 96.1 \\
		Pill  & 70.0  & 69.1  & 75.7  & 78.5  & 75.7  & 94.5  & 94.5  & 88.8 \\
		Screw & 59.1  & 79.0  & 60.9  & 78.5  & 65.4  & 70.0  & 79.0  & 96.3 \\
		Toothbrush & 59.2  & 65.6  & 94.2  & 91.1  & 94.2  & 99.2  & 99.2  & 98.6 \\
		Transistor & 52.0  & 50.5 & 77.3  & 59.0  & 79.7  & 91.9  & 91.9  & 91.1 \\
		Zipper & 69.3  & 88.3  & 51.2  & 98.0  & 85.5  & 98.7  & 98.7  & 95.1 \\
		\hline
		Carpet & 36.4  & 62.4  & 69.1  & 48.0  & 90.4  & 98.4  & 98.4  & 92.9 \\
		Grid  & 35.1  & 57.6  & 56.6  & 38.5  & 65.5  & 89.6  & 89.6  & 84.0 \\
		Leather & 47.4  & 63.5  & 66.9  & 70.7  & 99.7  & 100.0 & 100.0 & 97.1 \\
		Tile  & 47.8  & 81.4  & 63.9  & 88.7  & 82.2  & 96.9  & 96.9  & 99.4 \\
		Wood  & 92.5  & 88.7  & 97.9  & 78.4  & 94.4  & 98.1  & 98.1  & 99.8 \\
		\hline
		\textbf{Average} & 55.8  & 57.3  & 72.6  & 72.9  & 83.7  & 94.5  & \textbf{95.1}  &94.7 \\
		    \toprule
	\end{tabular}%
	\label{tab:addlabel}%
\end{table*}%

\subsection{Few-shots learning experimental results}
\label{Sec:Fewshot_results}
\begin{figure*}[!h]
	\centering
	\includegraphics{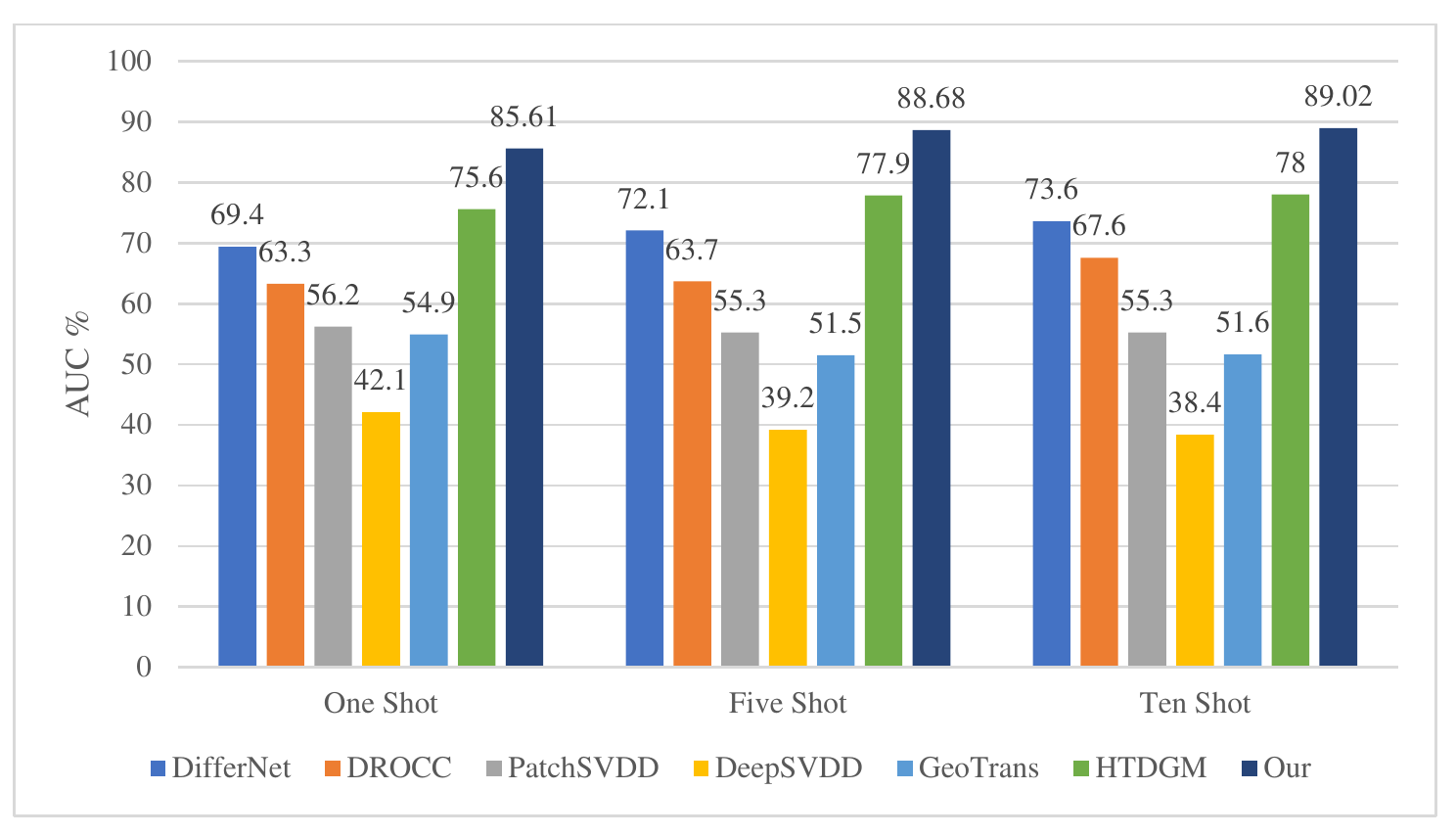}
	\caption{AUC in \% performance of the MVTec AD datase, for One shot, Five Shot, and Ten Shot settings.}
	\label{Fig:Mvtec_mean_fewshot}
\end{figure*}

\begin{figure}[!h]
	\includegraphics[width=0.48\textwidth]{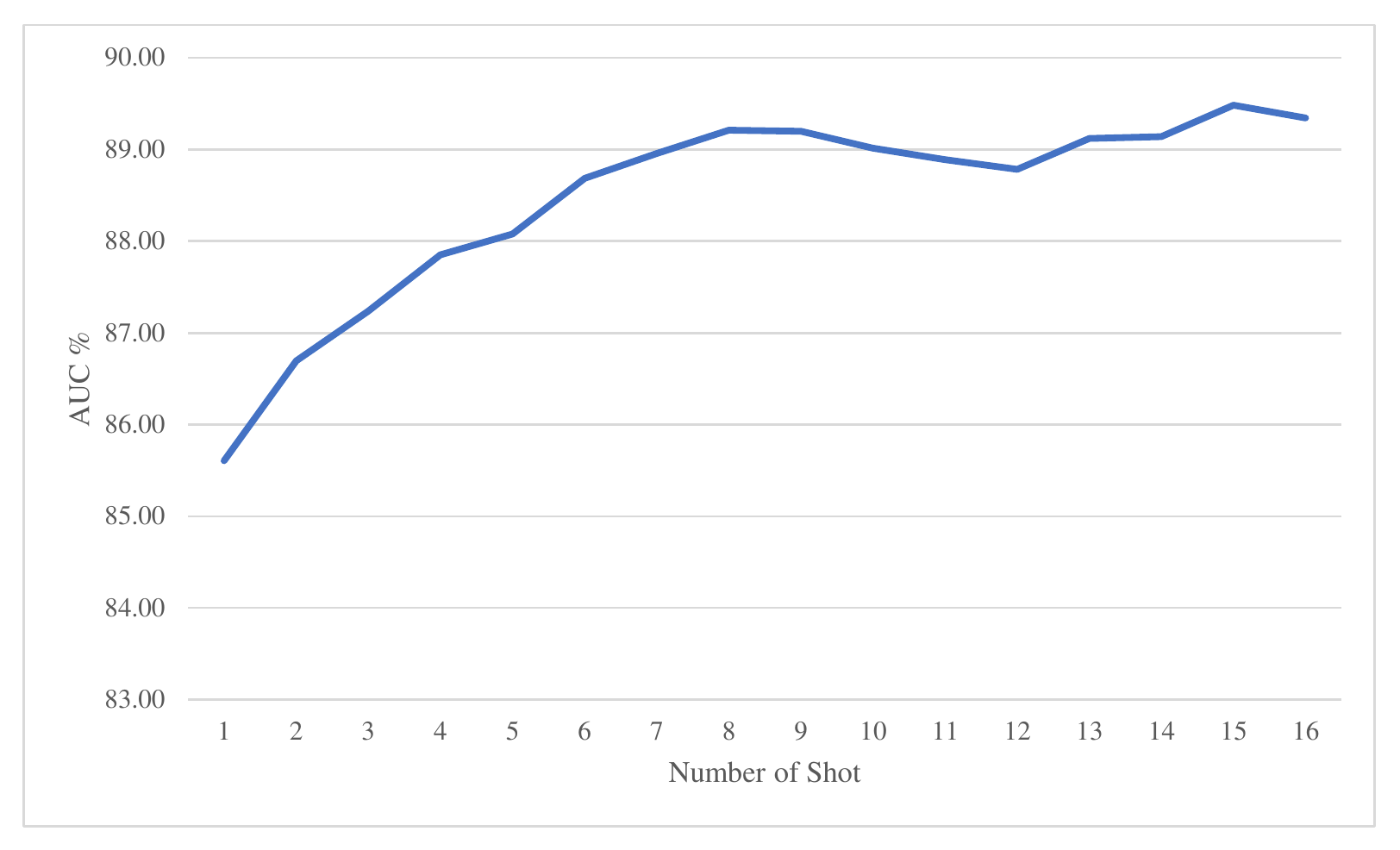}
	\caption{The effect of increasing the number of samples on the average AUC in \% of our proposed approach for the MVTec AD dataset.}
	\label{Fig:Mvtec_mean_overshots}
\end{figure}

\begin{figure}[!h]
	\includegraphics[width=0.48\textwidth]{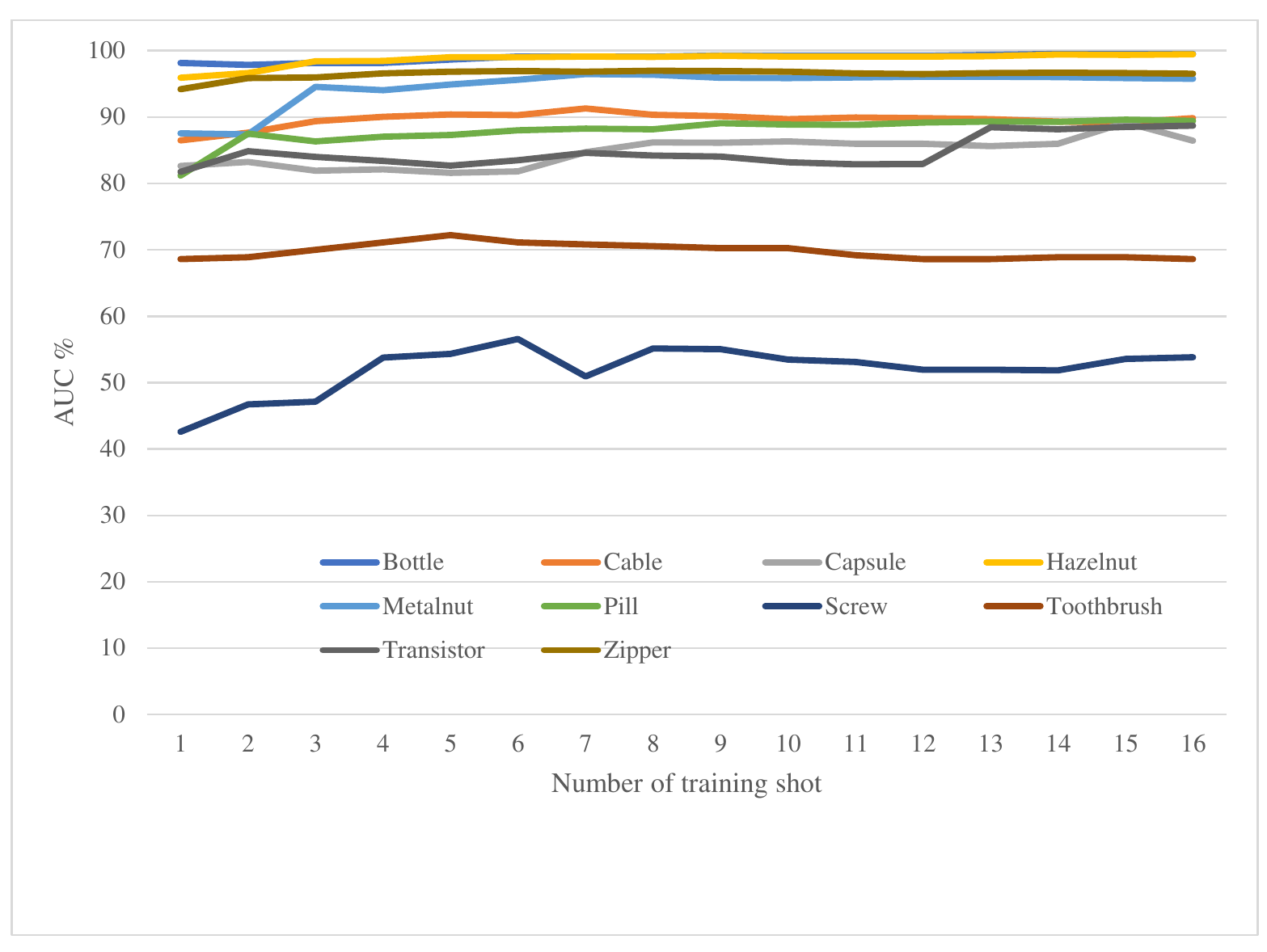}
	\caption{The effect of increasing the number of samples on AUC in \% of our proposed approach for 10 objects of MVTec AD dataset.}
	\label{Fig:Mvtec_objects_overshots}
\end{figure}

\begin{figure}[!h]
	\includegraphics[width=0.48\textwidth]{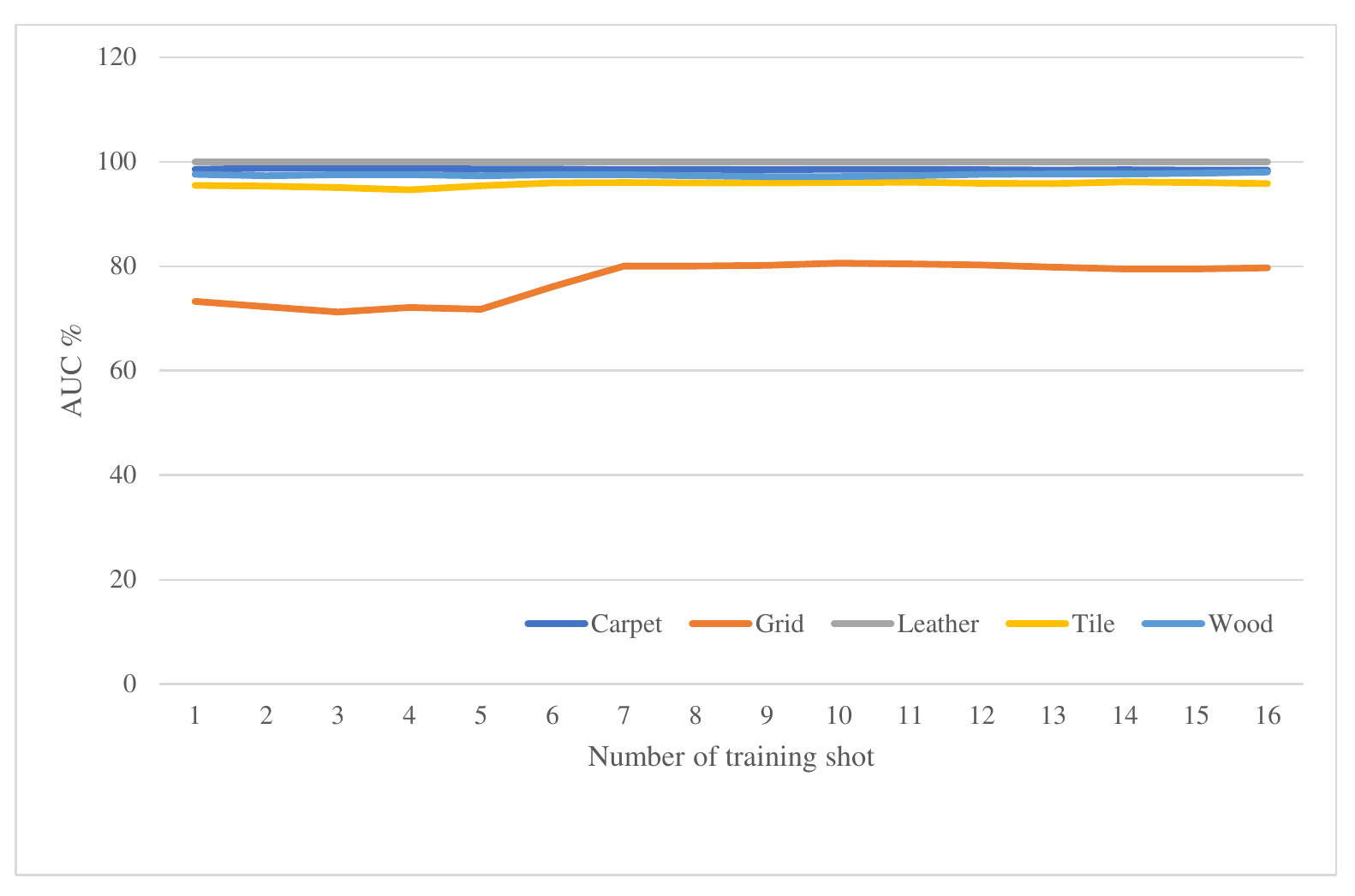}
	\caption{The effect of increasing the number of samples on AUC in \% of our proposed approach for 5 textures of MVTec AD dataset.}
	\label{Fig:Mvtec_textures_overshots}
\end{figure}

In this section, we demonstrate the effectiveness of our proposed approach via few-shots learning. Few shot learning studies the use of limited supervision for classification tasks and has been widely studied~\cite{chen2019closer,wang2020generalizing}. Other works consider a limited number of samples from anomalous classes~\cite{pang2018learning,pang2019deep}. The most related work proposed by Sheynin~\andothers~\cite{sheynin2021hierarchical} proposes to use a limited number of normal samples for anomaly detection~\cite{sheynin2021hierarchical}. We compare our approach with the baseline methods in~\cite{sheynin2021hierarchical}. The comparison includes DifferNet~\cite{rudolph2021same}, GeoTrans~\cite{geotrans}, and GOAD~\cite{bergman2020classification} a modified version of GeoTrans that modify the anomaly score. DeepSVDD~\cite{ruff2018deep} uses a deep version of SVDD algorithm of the pre-trained deep features. PatchSVDD~\cite{defard2020padim} use self supervised learning and extend DeepSVDD to patch-based method. DROCC~\cite{goyal2020drocc} approach train classier to distinguish between normal samples and their perturbed versions generated adversarially. Finally, we compare our method with HTDGM~\cite{sheynin2021hierarchical} that use self supervised learning of the multis-scale hierarchical transformation of the generative model. Experiments for MVTec AD dataset is shown in Fig.~\ref{Fig:Mvtec_mean_fewshot}. We follow the same evaluation protocol and also the same D2-Net for feature extraction Fig.~\ref{Fig:Mvtec_mean_fewshot} shows that the proposed approach outperforms all other methods in all few shot settings. Fig.~\ref{Fig:Mvtec_mean_overshots} shows the mean AUC of our proposed model for MVTec AD dataset over different shots. More details about the AUC accuracy for each category over the 16 shot are shown in Figs.~\ref{Fig:Mvtec_objects_overshots} and ~\ref{Fig:Mvtec_textures_overshots}.


\section{Conclusion}
\label{Sec:conclusion}
This paper proposes the use of transfer learning of local features as an efficient alternative to global features for defect detection. The use of deep local features has been limited to SLAM or SFM applications and has not been used for anomaly/defect detection in the literature. The pre-trained local features network returns a set of point pattern/keypoints with their descriptors, and these features are well-known to be robust against viewpoints and lighting changes. We propose to model these descriptors as random finite sets and model the similarity of normal distribution based on RFS statistics. We propose to use RFS energy instead of using RFS log-likelihood as an anomaly score. The proposed IID Possion RFS energy includes learning the Mahalanobis distance parameters and Poisson intensity of the cardinality distribution. To avoid further preprocessing of features and loss of information due to PCA, we choose to learn the parameters of the Mahalanobis distance on the high dimension space. Experimental results on the MVTec AD dataset, a challenging surface defect detection dataset, show that our proposed approach has outstanding performance compared to the state-of-the-arts approaches. The proposed approach outperforms other methods in 11 out 15 objects/textures. By using the different local features extraction shown in the ablation study, we can outperform the state-of-art on MVTec AD dataset. Another experiment has been conducted on MVTec AD using a few-shots learning setting, in which only a few samples are used in the training phase.  In the few-shots setting, the approach RFS energy outperforms the state-of-art methods in 1, 5, 10 shots.


%



\section*{Acknowledgment}
This work was supported by the Australian Research Council (the ARC) via the Project Linkage grant LP160101081.

\ifCLASSOPTIONcaptionsoff
  \newpage
\fi



\bibliographystyle{IEEEtran}
\bibliography{references}

\begin{thebibliography}{10}
\providecommand{\url}[1]{#1}
\csname url@samestyle\endcsname
\providecommand{\newblock}{\relax}
\providecommand{\bibinfo}[2]{#2}
\providecommand{\BIBentrySTDinterwordspacing}{\spaceskip=0pt\relax}
\providecommand{\BIBentryALTinterwordstretchfactor}{4}
\providecommand{\BIBentryALTinterwordspacing}{\spaceskip=\fontdimen2\font plus
\BIBentryALTinterwordstretchfactor\fontdimen3\font minus
  \fontdimen4\font\relax}
\providecommand{\BIBforeignlanguage}[2]{{%
\expandafter\ifx\csname l@#1\endcsname\relax
\typeout{** WARNING: IEEEtran.bst: No hyphenation pattern has been}%
\typeout{** loaded for the language `#1'. Using the pattern for}%
\typeout{** the default language instead.}%
\else
\language=\csname l@#1\endcsname
\fi
#2}}
\providecommand{\BIBdecl}{\relax}
\BIBdecl

\bibitem{ngan2011automated}
H.~Y. Ngan, G.~K. Pang, and N.~H. Yung, ``Automated fabric defect detection—a
  review,'' \emph{Image and vision computing}, vol.~29, no.~7, pp. 442--458,
  2011.

\bibitem{gao2021review}
Y.~Gao, X.~Li, X.~V. Wang, L.~Wang, and L.~Gao, ``A review on recent advances
  in vision-based defect recognition towards industrial intelligence,''
  \emph{Journal of Manufacturing Systems}, 2021.

\bibitem{neogi2014review}
N.~Neogi, D.~K. Mohanta, and P.~K. Dutta, ``Review of vision-based steel
  surface inspection systems,'' \emph{EURASIP Journal on Image and Video
  Processing}, vol. 2014, no.~1, pp. 1--19, 2014.

\bibitem{wang2018simple}
H.~Wang, J.~Zhang, Y.~Tian, H.~Chen, H.~Sun, and K.~Liu, ``A simple guidance
  template-based defect detection method for strip steel surfaces,'' \emph{IEEE
  Transactions on Industrial Informatics}, vol.~15, no.~5, pp. 2798--2809,
  2018.

\bibitem{he2019fully}
T.~He, Y.~Liu, C.~Xu, X.~Zhou, Z.~Hu, and J.~Fan, ``A fully convolutional
  neural network for wood defect location and identification,'' \emph{IEEE
  Access}, vol.~7, pp. 123\,453--123\,462, 2019.

\bibitem{hanzaei2017automatic}
S.~H. Hanzaei, A.~Afshar, and F.~Barazandeh, ``Automatic detection and
  classification of the ceramic tiles’ surface defects,'' \emph{Pattern
  Recognition}, vol.~66, pp. 174--189, 2017.

\bibitem{shen2013automated}
H.-K. Shen, P.-H. Chen, and L.-M. Chang, ``Automated steel bridge coating rust
  defect recognition method based on color and texture feature,''
  \emph{Automation in Construction}, vol.~31, pp. 338--356, 2013.

\bibitem{elbehiery2005visual}
H.~M. Elbehiery, A.~A. Hefnawy, and M.~T. Elewa, ``Visual inspection for fired
  ceramic tile's surface defects using wavelet analysis,'' \emph{GVIP (05)},
  no.~V2, pp. 1--8, 2005.

\bibitem{jing2013fabric}
J.~Jing, H.~Zhang, J.~Wang, P.~Li, and J.~Jia, ``Fabric defect detection using
  gabor filters and defect classification based on lbp and tamura method,''
  \emph{Journal of the Textile Institute}, vol. 104, no.~1, pp. 18--27, 2013.

\bibitem{lowe2004distinctive}
D.~G. Lowe, ``Distinctive image features from scale-invariant keypoints,''
  \emph{International journal of computer vision}, vol.~60, no.~2, pp. 91--110,
  2004.

\bibitem{shumin2011adaboost}
D.~Shumin, L.~Zhoufeng, and L.~Chunlei, ``Adaboost learning for fabric defect
  detection based on hog and svm,'' in \emph{2011 International conference on
  multimedia technology}.\hskip 1em plus 0.5em minus 0.4em\relax IEEE, 2011,
  pp. 2903--2906.

\bibitem{wang2018deep}
J.~Wang, Y.~Ma, L.~Zhang, R.~X. Gao, and D.~Wu, ``Deep learning for smart
  manufacturing: Methods and applications,'' \emph{Journal of Manufacturing
  Systems}, vol.~48, pp. 144--156, 2018.

\bibitem{dai2020soldering}
W.~Dai, A.~Mujeeb, M.~Erdt, and A.~Sourin, ``Soldering defect detection in
  automatic optical inspection,'' \emph{Advanced Engineering Informatics},
  vol.~43, p. 101004, 2020.

\bibitem{yin2020deep}
X.~Yin, Y.~Chen, A.~Bouferguene, H.~Zaman, M.~Al-Hussein, and L.~Kurach, ``A
  deep learning-based framework for an automated defect detection system for
  sewer pipes,'' \emph{Automation in Construction}, vol. 109, p. 102967, 2020.

\bibitem{yu2017fully}
Z.~Yu, X.~Wu, and X.~Gu, ``Fully convolutional networks for surface defect
  inspection in industrial environment,'' in \emph{International conference on
  computer vision systems}.\hskip 1em plus 0.5em minus 0.4em\relax Springer,
  2017, pp. 417--426.

\bibitem{wu2017surface}
X.~Wu, K.~Cao, and X.~Gu, ``A surface defect detection based on convolutional
  neural network,'' in \emph{International Conference on Computer Vision
  Systems}.\hskip 1em plus 0.5em minus 0.4em\relax Springer, 2017, pp.
  185--194.

\bibitem{wang2019lednet}
Y.~Wang, Q.~Zhou, J.~Liu, J.~Xiong, G.~Gao, X.~Wu, and L.~J. Latecki, ``Lednet:
  A lightweight encoder-decoder network for real-time semantic segmentation,''
  in \emph{2019 IEEE International Conference on Image Processing
  (ICIP)}.\hskip 1em plus 0.5em minus 0.4em\relax IEEE, 2019, pp. 1860--1864.

\bibitem{hwang2019hexagan}
U.~Hwang, D.~Jung, and S.~Yoon, ``Hexagan: Generative adversarial nets for real
  world classification,'' \emph{arXiv preprint arXiv:1902.09913}, 2019.

\bibitem{oliver2018realistic}
A.~Oliver, A.~Odena, C.~A. Raffel, E.~D. Cubuk, and I.~Goodfellow, ``Realistic
  evaluation of deep semi-supervised learning algorithms,'' in \emph{Advances
  in Neural Information Processing Systems}, 2018, pp. 3235--3246.

\bibitem{yoon2018gain}
J.~Yoon, J.~Jordon, and M.~Van Der~Schaar, ``Gain: Missing data imputation
  using generative adversarial nets,'' \emph{arXiv preprint arXiv:1806.02920},
  2018.

\bibitem{chandola2009anomaly}
V.~Chandola, A.~Banerjee, and V.~Kumar, ``Anomaly detection: A survey,''
  \emph{ACM computing surveys (CSUR)}, vol.~41, no.~3, pp. 1--58, 2009.

\bibitem{bergmann2019mvtec}
P.~Bergmann, M.~Fauser, D.~Sattlegger, and C.~Steger, ``Mvtec ad--a
  comprehensive real-world dataset for unsupervised anomaly detection,'' in
  \emph{Proceedings of the IEEE Conference on Computer Vision and Pattern
  Recognition}, 2019, pp. 9592--9600.

\bibitem{baur2018deep}
C.~Baur, B.~Wiestler, S.~Albarqouni, and N.~Navab, ``Deep autoencoding models
  for unsupervised anomaly segmentation in brain mr images,'' in
  \emph{International MICCAI Brainlesion Workshop}.\hskip 1em plus 0.5em minus
  0.4em\relax Springer, 2018, pp. 161--169.

\bibitem{rudolph2021same}
M.~Rudolph, B.~Wandt, and B.~Rosenhahn, ``Same same but differnet:
  Semi-supervised defect detection with normalizing flows,'' in
  \emph{Proceedings of the IEEE/CVF Winter Conference on Applications of
  Computer Vision}, 2021, pp. 1907--1916.

\bibitem{cao2020unifying}
B.~Cao, A.~Araujo, and J.~Sim, ``Unifying deep local and global features for
  image search,'' in \emph{European Conference on Computer Vision}.\hskip 1em
  plus 0.5em minus 0.4em\relax Springer, 2020, pp. 726--743.

\bibitem{tang2019gcnv2}
J.~Tang, L.~Ericson, J.~Folkesson, and P.~Jensfelt, ``Gcnv2: Efficient
  correspondence prediction for real-time slam,'' \emph{IEEE Robotics and
  Automation Letters}, vol.~4, no.~4, pp. 3505--3512, 2019.

\bibitem{schonberger2018semantic}
J.~L. Sch{\"o}nberger, M.~Pollefeys, A.~Geiger, and T.~Sattler, ``Semantic
  visual localization,'' in \emph{Proceedings of the IEEE conference on
  computer vision and pattern recognition}, 2018, pp. 6896--6906.

\bibitem{goodfellow2014generative}
I.~J. Goodfellow, J.~Pouget-Abadie, M.~Mirza, B.~Xu, D.~Warde-Farley, S.~Ozair,
  A.~Courville, and Y.~Bengio, ``Generative adversarial networks,'' \emph{arXiv
  preprint arXiv:1406.2661}, 2014.

\bibitem{kingma2013auto}
D.~P. Kingma and M.~Welling, ``Auto-encoding variational bayes,'' \emph{arXiv
  preprint arXiv:1312.6114}, 2013.

\bibitem{rudolph2019structuring}
M.~Rudolph, B.~Wandt, and B.~Rosenhahn, ``Structuring autoencoders,'' in
  \emph{Proceedings of the IEEE/CVF International Conference on Computer Vision
  Workshops}, 2019, pp. 0--0.

\bibitem{lecun1989generalization}
Y.~LeCun \emph{et~al.}, ``Generalization and network design strategies,''
  \emph{Connectionism in perspective}, vol.~19, pp. 143--155, 1989.

\bibitem{chen2005simultaneous}
D.~Chen, X.~Shao, B.~Hu, and Q.~Su, ``Simultaneous wavelength selection and
  outlier detection in multivariate regression of near-infrared spectra,''
  \emph{Analytical Sciences}, vol.~21, no.~2, pp. 161--166, 2005.

\bibitem{zhai2016deep}
S.~Zhai, Y.~Cheng, W.~Lu, and Z.~Zhang, ``Deep structured energy based models
  for anomaly detection,'' in \emph{International Conference on Machine
  Learning}.\hskip 1em plus 0.5em minus 0.4em\relax PMLR, 2016, pp. 1100--1109.

\bibitem{schlegl2019f}
T.~Schlegl, P.~Seeb{\"o}ck, S.~M. Waldstein, G.~Langs, and U.~Schmidt-Erfurth,
  ``f-anogan: Fast unsupervised anomaly detection with generative adversarial
  networks,'' \emph{Medical image analysis}, vol.~54, pp. 30--44, 2019.

\bibitem{akcay2018ganomaly}
S.~Akcay, A.~Atapour-Abarghouei, and T.~P. Breckon, ``Ganomaly: Semi-supervised
  anomaly detection via adversarial training,'' in \emph{Asian conference on
  computer vision}.\hskip 1em plus 0.5em minus 0.4em\relax Springer, 2018, pp.
  622--637.

\bibitem{donahuedeep}
J.~Donahue, Y.~Jia, O.~Vinyals, J.~Hoffman, N.~Zhang, E.~Tzeng, and T.~Darrell,
  ``A deep convolutional activation feature for generic visual recognition,''
  \emph{UC Berkeley \& ICSI, Berkeley, CA, USA}.

\bibitem{andrews2016transfer}
J.~Andrews, T.~Tanay, E.~J. Morton, and L.~D. Griffin, ``Transfer
  representation-learning for anomaly detection.''\hskip 1em plus 0.5em minus
  0.4em\relax JMLR, 2016.

\bibitem{simonyan2014very}
K.~Simonyan and A.~Zisserman, ``Very deep convolutional networks for
  large-scale image recognition,'' \emph{arXiv preprint arXiv:1409.1556}, 2014.

\bibitem{nazare2018pre}
T.~S. Nazare, R.~F. de~Mello, and M.~A. Ponti, ``Are pre-trained cnns good
  feature extractors for anomaly detection in surveillance videos?''
  \emph{arXiv preprint arXiv:1811.08495}, 2018.

\bibitem{sabokrou2018deep}
M.~Sabokrou, M.~Fayyaz, M.~Fathy, Z.~Moayed, and R.~Klette, ``Deep-anomaly:
  Fully convolutional neural network for fast anomaly detection in crowded
  scenes,'' \emph{Computer Vision and Image Understanding}, vol. 172, pp.
  88--97, 2018.

\bibitem{teichmann2019detect}
M.~Teichmann, A.~Araujo, M.~Zhu, and J.~Sim, ``Detect-to-retrieve: Efficient
  regional aggregation for image search,'' in \emph{Proceedings of the IEEE/CVF
  Conference on Computer Vision and Pattern Recognition}, 2019, pp. 5109--5118.

\bibitem{schonberger2016structure}
J.~L. Schonberger and J.-M. Frahm, ``Structure-from-motion revisited,'' in
  \emph{Proceedings of the IEEE conference on computer vision and pattern
  recognition}, 2016, pp. 4104--4113.

\bibitem{sattler2019understanding}
T.~Sattler, Q.~Zhou, M.~Pollefeys, and L.~Leal-Taixe, ``Understanding the
  limitations of cnn-based absolute camera pose regression,'' in
  \emph{Proceedings of the IEEE/CVF Conference on Computer Vision and Pattern
  Recognition}, 2019, pp. 3302--3312.

\bibitem{busam2018markerless}
B.~Busam, P.~Ruhkamp, S.~Virga, B.~Lentes, J.~Rackerseder, N.~Navab, and
  C.~Hennersperger, ``Markerless inside-out tracking for 3d ultrasound
  compounding,'' in \emph{Simulation, Image Processing, and Ultrasound Systems
  for Assisted Diagnosis and Navigation}.\hskip 1em plus 0.5em minus
  0.4em\relax Springer, 2018, pp. 56--64.

\bibitem{harris1988combined}
C.~G. Harris, M.~Stephens \emph{et~al.}, ``A combined corner and edge
  detector.'' in \emph{Alvey vision conference}, vol.~15, no.~50.\hskip 1em
  plus 0.5em minus 0.4em\relax Citeseer, 1988, pp. 10--5244.

\bibitem{beaudet1978rotationally}
P.~R. Beaudet, ``Rotationally invariant image operators,'' in \emph{Proc. 4th
  Int. Joint Conf. Pattern Recog, Tokyo, Japan, 1978}, 1978.

\bibitem{mikolajczyk2004scale}
K.~Mikolajczyk and C.~Schmid, ``Scale \& affine invariant interest point
  detectors,'' \emph{International journal of computer vision}, vol.~60, no.~1,
  pp. 63--86, 2004.

\bibitem{tuytelaars2008local}
T.~Tuytelaars and K.~Mikolajczyk, \emph{Local invariant feature detectors: a
  survey}.\hskip 1em plus 0.5em minus 0.4em\relax Now Publishers Inc, 2008.

\bibitem{bay2008speeded}
H.~Bay, A.~Ess, T.~Tuytelaars, and L.~Van~Gool, ``Speeded-up robust features
  (surf),'' \emph{Computer vision and image understanding}, vol. 110, no.~3,
  pp. 346--359, 2008.

\bibitem{alcantarilla2011fast}
P.~F. Alcantarilla and T.~Solutions, ``Fast explicit diffusion for accelerated
  features in nonlinear scale spaces,'' \emph{IEEE Trans. Patt. Anal. Mach.
  Intell}, vol.~34, no.~7, pp. 1281--1298, 2011.

\bibitem{matas2004robust}
J.~Matas, O.~Chum, M.~Urban, and T.~Pajdla, ``Robust wide-baseline stereo from
  maximally stable extremal regions,'' \emph{Image and vision computing},
  vol.~22, no.~10, pp. 761--767, 2004.

\bibitem{rosten2006machine}
E.~Rosten and T.~Drummond, ``Machine learning for high-speed corner
  detection,'' in \emph{European conference on computer vision}.\hskip 1em plus
  0.5em minus 0.4em\relax Springer, 2006, pp. 430--443.

\bibitem{leutenegger2011brisk}
S.~Leutenegger, M.~Chli, and R.~Y. Siegwart, ``Brisk: Binary robust invariant
  scalable keypoints,'' in \emph{2011 International conference on computer
  vision}.\hskip 1em plus 0.5em minus 0.4em\relax Ieee, 2011, pp. 2548--2555.

\bibitem{rublee2011orb}
E.~Rublee, V.~Rabaud, K.~Konolige, and G.~Bradski, ``Orb: An efficient
  alternative to sift or surf,'' in \emph{2011 International conference on
  computer vision}.\hskip 1em plus 0.5em minus 0.4em\relax Ieee, 2011, pp.
  2564--2571.

\bibitem{verdie2015tilde}
Y.~Verdie, K.~Yi, P.~Fua, and V.~Lepetit, ``Tilde: A temporally invariant
  learned detector,'' in \emph{Proceedings of the IEEE Conference on Computer
  Vision and Pattern Recognition}, 2015, pp. 5279--5288.

\bibitem{lenc2016learning}
K.~Lenc and A.~Vedaldi, ``Learning covariant feature detectors,'' in
  \emph{European conference on computer vision}.\hskip 1em plus 0.5em minus
  0.4em\relax Springer, 2016, pp. 100--117.

\bibitem{zhang2017learning}
X.~Zhang, F.~X. Yu, S.~Karaman, and S.-F. Chang, ``Learning discriminative and
  transformation covariant local feature detectors,'' in \emph{Proceedings of
  the IEEE conference on computer vision and pattern recognition}, 2017, pp.
  6818--6826.

\bibitem{barroso2019key}
A.~Barroso-Laguna, E.~Riba, D.~Ponsa, and K.~Mikolajczyk, ``Key. net: Keypoint
  detection by handcrafted and learned cnn filters,'' in \emph{Proceedings of
  the IEEE/CVF International Conference on Computer Vision}, 2019, pp.
  5836--5844.

\bibitem{detone2017toward}
D.~DeTone, T.~Malisiewicz, and A.~Rabinovich, ``Toward geometric deep slam,''
  \emph{arXiv preprint arXiv:1707.07410}, 2017.

\bibitem{detone2018superpoint}
------, ``Superpoint: Self-supervised interest point detection and
  description,'' in \emph{Proceedings of the IEEE Conference on Computer Vision
  and Pattern Recognition Workshops}, 2018, pp. 224--236.

\bibitem{yi2016lift}
K.~M. Yi, E.~Trulls, V.~Lepetit, and P.~Fua, ``Lift: Learned invariant feature
  transform,'' in \emph{European Conference on Computer Vision}.\hskip 1em plus
  0.5em minus 0.4em\relax Springer, 2016, pp. 467--483.

\bibitem{ono2018lf}
Y.~Ono, E.~Trulls, P.~Fua, and K.~M. Yi, ``Lf-net: learning local features from
  images,'' in \emph{Advances in Neural Information Processing Systems}, 2018,
  pp. 6237--6247.

\bibitem{revaud2019r2d2}
J.~Revaud, P.~Weinzaepfel, C.~De~Souza, N.~Pion, G.~Csurka, Y.~Cabon, and
  M.~Humenberger, ``R2d2: Repeatable and reliable detector and descriptor,''
  \emph{arXiv preprint arXiv:1906.06195}, 2019.

\bibitem{tian2020d2d}
Y.~Tian, V.~Balntas, T.~Ng, A.~Barroso-Laguna, Y.~Demiris, and K.~Mikolajczyk,
  ``D2d: Keypoint extraction with describe to detect approach,'' in
  \emph{Proceedings of the Asian Conference on Computer Vision}, 2020.

\bibitem{dusmanu2019d2}
M.~Dusmanu, I.~Rocco, T.~Pajdla, M.~Pollefeys, J.~Sivic, A.~Torii, and
  T.~Sattler, ``D2-net: A trainable cnn for joint description and detection of
  local features,'' in \emph{Proceedings of the IEEE Conference on Computer
  Vision and Pattern Recognition}, 2019, pp. 8092--8101.

\bibitem{chiu2013stochastic}
S.~N. Chiu, D.~Stoyan, W.~S. Kendall, and J.~Mecke, \emph{Stochastic geometry
  and its applications}.\hskip 1em plus 0.5em minus 0.4em\relax John Wiley \&
  Sons, 2013.

\bibitem{vo2018model}
B.-N. Vo, N.~Dam, D.~Phung, Q.~N. Tran, and B.-T. Vo, ``Model-based learning
  for point pattern data,'' \emph{Pattern Recognition}, vol.~84, pp. 136--151,
  2018.

\bibitem{1261119Mahler}
R.~P.~S. {Mahler}, ``Multitarget bayes filtering via first-order multitarget
  moments,'' \emph{IEEE Transactions on Aerospace and Electronic Systems},
  vol.~39, no.~4, pp. 1152--1178, 2003.

\bibitem{Kamoona9074564}
A.~M. {Kamoona}, A.~K. {Gostar}, A.~{Bab-Hadiashar}, and R.~{Hoseinnezhad},
  ``Sparsity-based naive bayes approach for anomaly detection in real
  surveillance videos,'' in \emph{2019 International Conference on Control,
  Automation and Information Sciences (ICCAIS)}, 2019, pp. 1--6.

\bibitem{mahler2007statistical}
R.~P. Mahler, \emph{Statistical multisource-multitarget information
  fusion}.\hskip 1em plus 0.5em minus 0.4em\relax Artech House Norwood, MA,
  2007, vol. 685.

\bibitem{vo2008cardinality}
B.-T. Vo, B.-N. Vo, and A.~Cantoni, ``The cardinality balanced multi-target
  multi-bernoulli filter and its implementations,'' \emph{IEEE Transactions on
  Signal Processing}, vol.~57, no.~2, pp. 409--423, 2008.

\bibitem{papi2015generalized}
F.~Papi, B.-N. Vo, B.-T. Vo, C.~Fantacci, and M.~Beard, ``Generalized labeled
  multi-bernoulli approximation of multi-object densities,'' \emph{IEEE
  Transactions on Signal Processing}, vol.~63, no.~20, pp. 5487--5497, 2015.

\bibitem{Vo2018}
B.-N. Vo, N.~Dam, D.~Phung, Q.~N. Tran, and B.-T. Vo, ``{Model-based learning
  for point pattern data},'' \emph{Pattern Recognition}, vol.~84, pp. 136--151,
  dec 2018.

\bibitem{kamoona2019random}
A.~M. Kamoona, A.~K. Gostar, R.~Tennakoon, A.~Bab-Hadiashar, D.~Accadia,
  J.~Thorpe, and R.~Hoseinnezhad, ``Random finite set-based anomaly detection
  for safety monitoring in construction sites,'' \emph{IEEE Access}, vol.~7,
  pp. 105\,710--105\,720, 2019.

\bibitem{goodfellow2016deep}
I.~Goodfellow, Y.~Bengio, and A.~Courville, \emph{Deep learning}.\hskip 1em
  plus 0.5em minus 0.4em\relax MIT press, 2016.

\bibitem{zong2018deep}
B.~Zong, Q.~Song, M.~R. Min, W.~Cheng, C.~Lumezanu, D.~Cho, and H.~Chen, ``Deep
  autoencoding gaussian mixture model for unsupervised anomaly detection,'' in
  \emph{International conference on learning representations}, 2018.

\bibitem{grathwohl2019your}
W.~Grathwohl, K.-C. Wang, J.-H. Jacobsen, D.~Duvenaud, M.~Norouzi, and
  K.~Swersky, ``Your classifier is secretly an energy based model and you
  should treat it like one,'' \emph{arXiv preprint arXiv:1912.03263}, 2019.

\bibitem{liu2020energy}
W.~Liu, X.~Wang, J.~D. Owens, and Y.~Li, ``Energy-based out-of-distribution
  detection,'' \emph{arXiv preprint arXiv:2010.03759}, 2020.

\bibitem{mahalanobis1936generalized}
P.~C. Mahalanobis, ``On the generalized distance in statistics.''\hskip 1em
  plus 0.5em minus 0.4em\relax National Institute of Science of India, 1936.

\bibitem{christiansen2016deepanomaly}
P.~Christiansen, L.~N. Nielsen, K.~A. Steen, R.~N. J{\o}rgensen, and
  H.~Karstoft, ``Deepanomaly: Combining background subtraction and deep
  learning for detecting obstacles and anomalies in an agricultural field,''
  \emph{Sensors}, vol.~16, no.~11, p. 1904, 2016.

\bibitem{rippel2021modeling}
O.~Rippel, P.~Mertens, and D.~Merhof, ``Modeling the distribution of normal
  data in pre-trained deep features for anomaly detection,'' in \emph{2020 25th
  International Conference on Pattern Recognition (ICPR)}.\hskip 1em plus 0.5em
  minus 0.4em\relax IEEE, 2021, pp. 6726--6733.

\bibitem{lee2018simple}
K.~Lee, K.~Lee, H.~Lee, and J.~Shin, ``A simple unified framework for detecting
  out-of-distribution samples and adversarial attacks,'' \emph{Advances in
  neural information processing systems}, vol.~31, 2018.

\bibitem{ledoit2004well}
O.~Ledoit and M.~Wolf, ``A well-conditioned estimator for large-dimensional
  covariance matrices,'' \emph{Journal of multivariate analysis}, vol.~88,
  no.~2, pp. 365--411, 2004.

\bibitem{ganomaly}
\BIBentryALTinterwordspacing
S.~Akcay, A.~A. Abarghouei, and T.~P. Breckon, ``Ganomaly: Semi-supervised
  anomaly detection via adversarial training,'' \emph{CoRR}, vol.
  abs/1805.06725, 2018. [Online]. Available:
  \url{http://arxiv.org/abs/1805.06725}
\BIBentrySTDinterwordspacing

\bibitem{andrews}
J.~Andrews, T.~Tanay, E.~Morton, and L.~Griffin, ``Transfer
  representation-learning for anomaly detection.''\hskip 1em plus 0.5em minus
  0.4em\relax JMLR, 2016.

\bibitem{nazare}
T.~Nazare, R.~de~Mello, and M.~Ponti, ``Are pre-trained cnns good feature
  extractors for anomaly detection in surveillance videos?'' \emph{arXiv
  preprint arXiv:1811.08495}, 2018.

\bibitem{geotrans}
I.~Golan and R.~El-Yaniv, ``Deep anomaly detection using geometric
  transformations,'' in \emph{Advances in Neural Information Processing
  Systems}, 2018, pp. 9758--9769.

\bibitem{dsebm}
S.~Zhai, Y.~Cheng, W.~Lu, and Z.~Zhang, ``Deep structured energy based models
  for anomaly detection,'' in \emph{Proceedings of the 33rd International
  Conference on International Conference on Machine Learning-Volume 48}, 2016,
  pp. 1100--1109.

\bibitem{chen2019closer}
W.-Y. Chen, Y.-C. Liu, Z.~Kira, Y.-C.~F. Wang, and J.-B. Huang, ``A closer look
  at few-shot classification,'' \emph{arXiv preprint arXiv:1904.04232}, 2019.

\bibitem{wang2020generalizing}
Y.~Wang, Q.~Yao, J.~T. Kwok, and L.~M. Ni, ``Generalizing from a few examples:
  A survey on few-shot learning,'' \emph{ACM Computing Surveys (CSUR)},
  vol.~53, no.~3, pp. 1--34, 2020.

\bibitem{pang2018learning}
G.~Pang, L.~Cao, L.~Chen, and H.~Liu, ``Learning representations of
  ultrahigh-dimensional data for random distance-based outlier detection,'' in
  \emph{Proceedings of the 24th ACM SIGKDD international conference on
  knowledge discovery \& data mining}, 2018, pp. 2041--2050.

\bibitem{pang2019deep}
G.~Pang, C.~Shen, and A.~van~den Hengel, ``Deep anomaly detection with
  deviation networks,'' in \emph{Proceedings of the 25th ACM SIGKDD
  international conference on knowledge discovery \& data mining}, 2019, pp.
  353--362.

\bibitem{sheynin2021hierarchical}
S.~Sheynin, S.~Benaim, and L.~Wolf, ``A hierarchical
  transformation-discriminating generative model for few shot anomaly
  detection,'' \emph{arXiv preprint arXiv:2104.14535}, 2021.

\bibitem{bergman2020classification}
L.~Bergman and Y.~Hoshen, ``Classification-based anomaly detection for general
  data,'' \emph{arXiv preprint arXiv:2005.02359}, 2020.

\bibitem{ruff2018deep}
L.~Ruff, R.~Vandermeulen, N.~Goernitz, L.~Deecke, S.~A. Siddiqui, A.~Binder,
  E.~M{\"u}ller, and M.~Kloft, ``Deep one-class classification,'' in
  \emph{International conference on machine learning}.\hskip 1em plus 0.5em
  minus 0.4em\relax PMLR, 2018, pp. 4393--4402.

\bibitem{defard2020padim}
T.~Defard, A.~Setkov, A.~Loesch, and R.~Audigier, ``Padim: a patch distribution
  modeling framework for anomaly detection and localization,'' \emph{arXiv
  preprint arXiv:2011.08785}, 2020.

\bibitem{goyal2020drocc}
S.~Goyal, A.~Raghunathan, M.~Jain, H.~V. Simhadri, and P.~Jain, ``Drocc: Deep
  robust one-class classification,'' in \emph{International Conference on
  Machine Learning}.\hskip 1em plus 0.5em minus 0.4em\relax PMLR, 2020, pp.
  3711--3721.

\end{thebibliography}
%

%

\begin{IEEEbiographynophoto}{Ammar Kamoona}
was awarded Master degree in electronic
and electrical engineering from Swinburne University of Technology,
Melbourne, Australia, 2016. He has been a student member of IEEE
since 2013, and the recipient of two certificates of excellence in RF
circuit design and Stochastic and Survival analysis from Swinburne
University, as well as a golden key certificate for being one of the top
achiever students. He worked as Assistant lecturer at Department of
Electrical Engineering, University of Kufa, Iraq from 2017 t0 2018.
Currently, Ammar is a PhD student at RMIT University of technology,
Melbourne, Australia. His current research interests include
Computer Vision, RFS filters, Robotics and Optimization, and FPGA
applications.
\end{IEEEbiographynophoto}

\begin{IEEEbiographynophoto}{Amirali Khodadadian Gostar}
received his BSc degree in Electrical
Engineering and MSc degree in Philosophy of Science, PhD degree in
Mechatronics Engineering from RMIT University. He is currently a
postdoctoral research fellow in School of Engineering at RMIT
University. His research interests include sensor management, data
fusion, and multitarget tracking.
\end{IEEEbiographynophoto}


\begin{IEEEbiographynophoto}{Alireza Bab-Hadiashar}
received his BSc and MEng in Mechanical
Engineering, then PhD in Robotics from Monash University. He has
held various positions in Monash University, Swinburne University of
Technology and RMIT University where he is currently a professor of
mechatronics and leads the intelligent automation research group.
His main area of research interest is intelligent automation in general,
and robust data fitting in machine vision, deep learning for detection
and identification, and robust data segmentation, in particular
\end{IEEEbiographynophoto}
\begin{IEEEbiographynophoto}{Reza Hoseinnezhad}
received his BSc, MSc, and PhD in Electrical
Engineering from University of Tehran, Iran, in 1994, 1996 and
2002, respectively. He has held various positions at University of
Tehran, Swinburne University of Technology, The University of
Melbourne, and RMIT University where he has worked since 2010
and is currently a Professor and Research Development Lead as
well as the Discipline Leader (Manufacturing \& Mechatronics) at
School of Engineering. His main areas of research interest are
statistical information fusion, random finite sets, multi-object
tracking, deep learning, and robust multi-structure data fitting in
computer vision.
\end{IEEEbiographynophoto}



\appendices
\newpage
\section{Samples of D2\_Net features for the MVTec AD dataset}
\centering
\includepdf[pages={1,2,3,4},fitpaper=True]{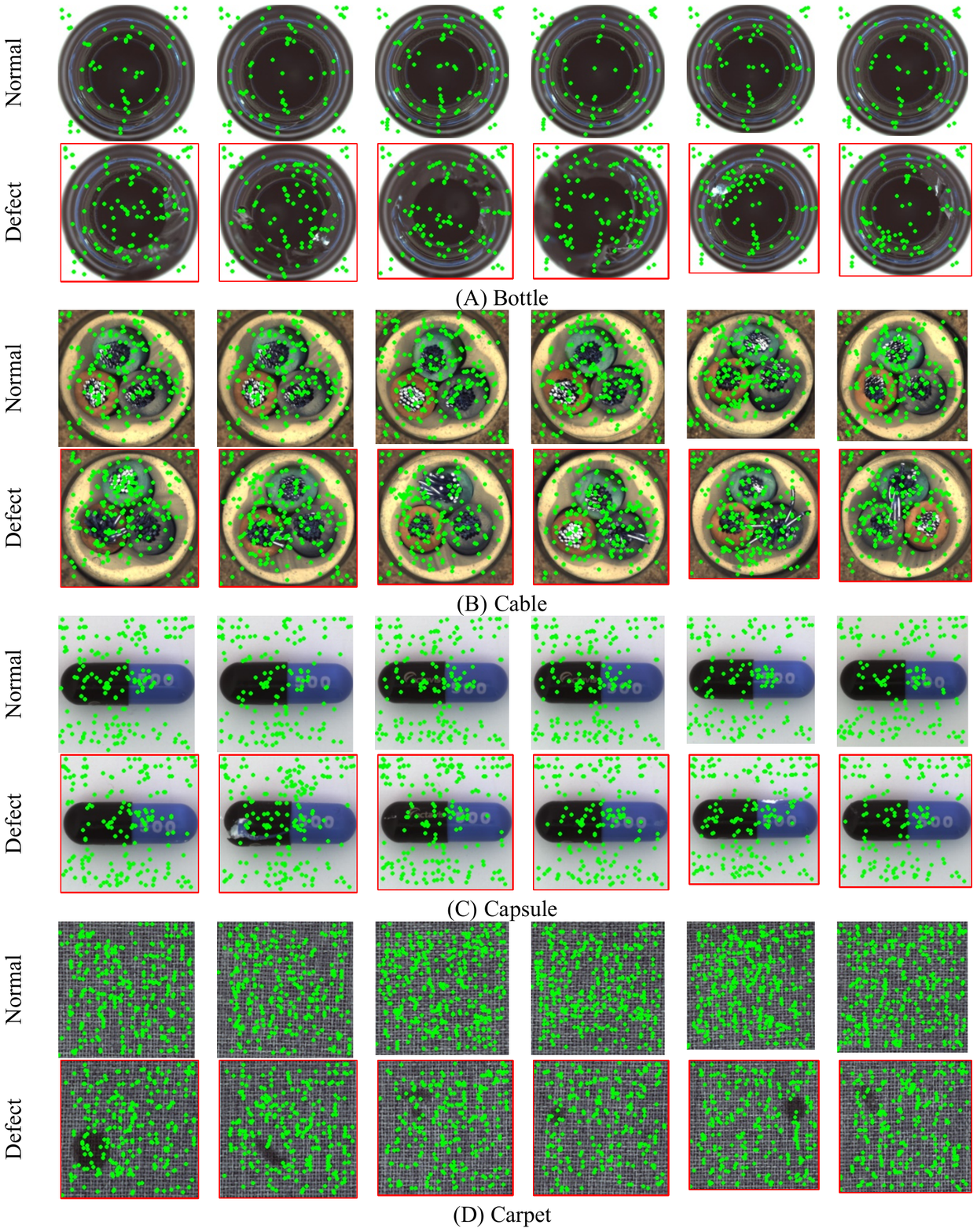}


\end{document}